\documentclass{2023Wu}
\usepackage{moreverb,url}
\usepackage[colorlinks,bookmarksopen,bookmarksnumbered,citecolor=red,urlcolor=red]{hyperref}
\usepackage{times}
\usepackage{moreverb,url}
\usepackage{subcaption}
\usepackage{amsfonts,amssymb,amsbsy,amsmath,amsthm}
\usepackage{xcolor}
\usepackage{graphicx,setspace}
\usepackage[mathlines, running]{lineno}
\usepackage{authblk}
\usepackage[compress, numbers]{natbib}
\usepackage{dsfont}
\usepackage[mathscr]{eucal}
\usepackage{listings}

\newcommand{\Chg}[1]{{#1}}
\newcommand{\Chgp}[1]{{#1}}
\newcommand{\xOf}[2]{{#1}_{{x,{#2}}}}
\newcommand{\xyOf}[2]{{#1}_{{xy,{#2}}}}
\newcommand{\yOf}[2]{{#1}_{{y,{#2}}}}
\newcommand{\zOf}[2]{{#1}_{{z,{#2}}}}
\newcommand{\Skw}{\mathrm{S}}
\newcommand{\Rot}{{\mathrm{R}_\theta}}
\newcommand{\q}{q}
\newcommand{\p}{p}
\newcommand{\w}{w}
\newcommand{\s}{\mathbf{s}}
\newcommand{\T}{\mathsf{T}}
\newcommand{\tw}{\Omega}

\newcommand{\HT}{\mathrm{H}}
\newcommand{\R}{\mathbb{R}}

\newcommand{\refFig}[1]{figure~\ref{fig:#1}}
\newcommand{\refEqn}[1]{eqn.~\ref{eqn:#1}}
\newcommand{\sgn}{\mathrm{sign}}
\newcommand{\CV}{viscous-Coulomb}

\begin{document}
\setlength\columnsep{25pt}
\title{Modeling multi-legged robot locomotion with slipping and its experimental validation}
\journalname{Multi-legged locomotion with slipping}

\author[1]{Ziyou Wu\footnote{wuziyou@umich.edu}}
\author[2,6]{Dan Zhao}
\author[1,3,4,5]{Shai Revzen}

\affil[1]{Robotics, University of Michigan}
\affil[2]{Mechanical Engineering, University of Michigan}
\affil[3]{Electrical Engineering and Computer Science, University of Michigan}
\affil[4]{Ecology and Evolutionary Biology, University of Michigan}
\affil[5]{Applied Physics Program, University of Michigan}
\affil[6]{Now at XPENG Robotics}

\maketitle
\tableofcontents

\begin{abstract}
Multi-legged robots with six or more legs are not in common use, despite designs with superior stability, maneuverability, and a low number of actuators being available for over 20 years.
This may be in part due to the difficulty in modeling multi-legged motion with slipping and producing reliable predictions of body velocity.
Here we present a detailed measurement of the foot contact forces in a hexapedal robot with multiple sliding contacts, and provide an algorithm for predicting these contact forces and the body velocity.
The algorithm relies on the recently published observation that even while slipping, multi-legged robots are principally kinematic, and employ a friction law ansatz that allows us to compute the shape-change to body-velocity connection and the foot contact forces.
\Chg{This results in the ability to simulate motion plans for a large number of contacts, each potentially with slipping.
Furthermore, in homogeneous environments, this kind of simulation can run in (parallel) logarithmic time of the planning horizon.}\\
\textbf{locomotion, friction, modeling, ground contact force}
\end{abstract}

\section{Introduction}
Most recent research in legged robots has focused on bipedal or quadrupedal robots, yet the vast majority of legged animal species use six or more legs,  are smaller, and therefore navigate a relatively much rougher terrain. 
Such ``multi-legged'' systems -- a term we refrain from using for quadrupeds and bipeds -- can exhibit complex trade-offs between loads on the legs, and move with substantial slipping at the feet.
\Chg{As the number of legs increases and the relative size of each leg becomes smaller, the locomotion of a multi-legged robot converges to that of a slithering of snake robot.
Slithering can be understood using geometric mechanics, and is ``principally kinematic'' in the sense of admitting equations of motion that do not require  momentum or velocity states.
Instead, the equations of motion can be expressed as a ``local connection'': a linear relationship between the rate of shape change $\dot r$ and the body velocity $v_b$: $v_b = A(r) \dot r$~(see e.g. \cite{Hatton-Choset2011, Geometric-Ostrowski1998}).  
It is natural to ask how this transition to momentum-free equations of motion occurs with increasing numbers of legs, and whether we can obtain a model for multi-legged locomotion without momentum.

Somewhat to our surprise, the minimal number of slipping point contacts at which momentum can be removed and a local connection describes the motion is three -- since with this number of contacts all DOF are subject either to rigid constraints or to sliding friction.
\cite{Bittner2018GOGO} presented a data-driven method to efficiently construct this local connection model for systems with no momentum.
The work in~\cite{Kvalheim2019perturbed} extended the no-momentum case to the quickly decaying momentum case which appears in high friction sliding contact.
Recently, \cite{Walk_slither_Zhao2022} showed that such local connection models are effective for multi-legged animals and robots with and without slipping, at the predicted level of three or more contacts.

The primary motivation of our work here is to demonstrate to the community that the body motion of multi-legged robots -- typically hexapods and octopods -- is much easier to model than is often assumed, because despite the complexity of additional contacts and a combinatorial number of contact states, the ensuing simplification that comes from momentum becoming ignorable more than compensates. Even if each leg ends in a multi-toed foot, leading to a total of a few dozen possible contact points, the simulation models remain eminently tractable. 

To achieve this goal we \textbf{(1)} present a fast algorithm for modeling multi-legged systems, that accurately estimates the body velocities needed for producing motion plans, and provides accurate foot contact force estimates.
We describe \textbf{(2)} a multi-legged robot that simultaneously measured all the ground contact forces produced while moving and possibly slipping, and \textbf{(3)} use this robot to test and validate the predictions of our model. 
We also \textbf{(4)} validate the model against other multi-legged robot datasets, and \textbf{(5)} demonstrate that it scales favorably with the number of legs and admits easy parallelization in comparison with a state-of-the-art mechanical simulator. 

Sufficiently high viscous friction has been known to produce a local connection since the seminal paper of \cite{shapere1989geometry}.
While several sources have compared predictions of viscous friction with Coulomb friction noting the similarity in results (most recently \cite{chong2022coordinating} supplemental figure 2), the two friction models are incompatible in an easily observable way. Coulomb friction models the fact that the traction generated by a leg depends on the normal force it exerts on the substrate whereas in viscous friction the traction depends only on velocity, making a viscous friction model ineffective except in cases of nearly identical leg loading.
To resolve this complication, we proposed the use of a \CV\ friction ansatz in \cite{revzen-2018-dw-tsml}, noting then the surprisingly good predictive quality of the ansatz-based model. 
This ansatz produces a local connection by construction, just as viscous friction does, while conventional Coulomb-friction based modeling approaches cannot produce such a linear relationship between shape velocity and body velocity. 
The current publication is the culmination of a multi-year effort to extend, understand, and validate those initial results (including, among other publications, \cite{Wu2019friction, zhao2021thesis, Revzen-Gravish-2021-allSpeeds-DW, Bittner-2020-dw, Revzen-Gravish-2021-bridging-DW, Walk_slither_Zhao2022}). 
}

\subsection{Multi-legged contact forces}

Much of the work on multi-legged robots has focused on hexpedal robots, starting with ~\cite{Gorinevsky1990hexapod, Carlton1987}.
Hexapods are appealing because they can have a tripod of supporting legs while moving another tripod into place for support on the next step. 
This static stability allows for the possibility of easier control, and usable motion even when the robot is slipping.
\Chg{Significant prior work was done on the hexpedal robots of the RHex family (from \cite{Rhex-Uluc2001} to \cite{roll-Rhex-Chang2022}), many variations of which have been built over the past 20 years.
Additional families of multi-legged robots include the RoACH robots of \cite{Roach-Zarrouk2015}, the Sprawl robots of \cite{iSprawl-Kim2006}, various multi-legged robots that used ``whegs'' from \cite{quinn2001insect} and e.g. \cite{1star-Zarrouk2015}, and several studies of multi-legged robots with larger numbers of legs such as \cite{Shinya2013centiped}, \cite{friction-Chong2022}, and \cite{multilegged-control-Chong2021}.}

\Chg{In a system with more than one point of contact,} every closed kinematic loop between body and ground can support an internal force that produces no net body acceleration.
This implies a sizable space of contact forces that unobservable from motion tracking \Chg{in a multilegged systems}, and cannot be measured using the commonly used approach of placing a force plate for the robot to move over \Chg{as in} \cite{Komsuoglu2009fp}.
This is because such a force plate only measures the total wrench applied to it -- not the interplay of all individual foot contact forces, which may trade off in various ways from step to step.
The authors of~\cite{Kao2019rhexFT} measured the individual ground contact forces of RHex by installing each leg with a 3d force sensor.

\Chg{\subsection{Motivating example}
\begin{figure}
\centering
\includegraphics[width=0.4\textwidth]{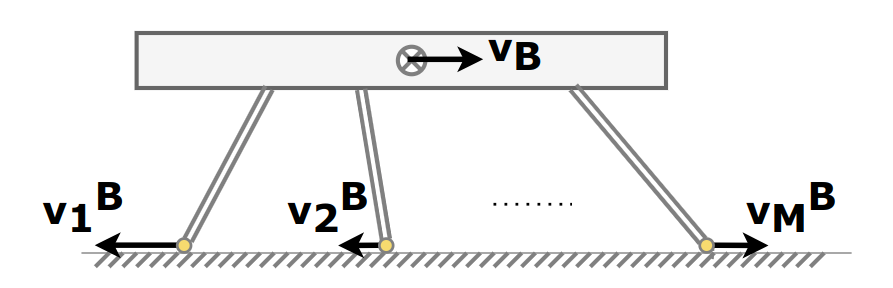}
\caption{\Chg{A 1-dimensional multilegged example. %
 Consider two or more equal-height legs, with identical friction properties moving at different horizontal velocities relative to a body constrained to move only in the horizontal direction. %
 The resulting body velocity under Coulomb friction with an odd number of legs is the median leg speed; with an even number of legs the answer is non-unique and any speed between the two median leg speeds will work.}}\label{fig:1d_ex} \
\end{figure}

\begin{figure}
\centering
\includegraphics[width=0.3\textwidth]{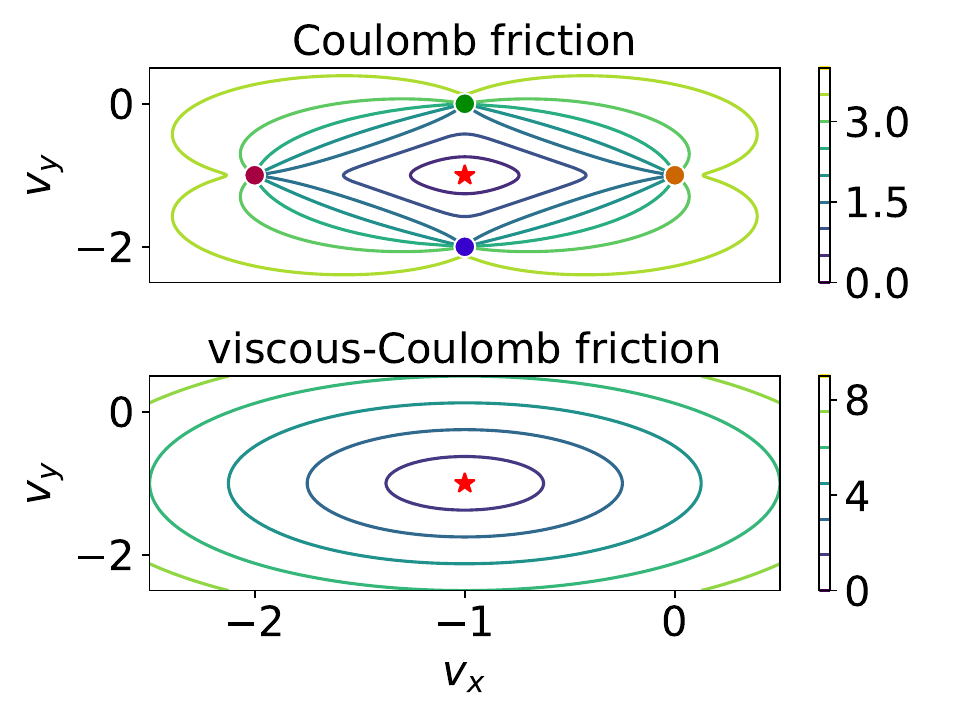}
\includegraphics[width=0.15\textwidth]{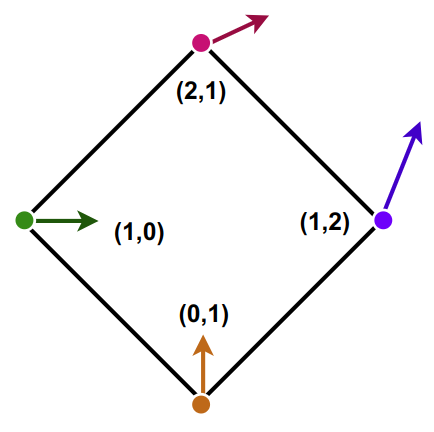}
\caption{\Chg{Visualization of a 2D case.\
Consider a 4-legged robot (right plot) whose feet move relative to the body with known velocities (arrows, right plot). \
We plotted contours of the magnitude of the total force on the body under the assumption that the body is moving at body velocity $(V_x, V_y)$ without rotating, both under Coulomb friction (top left) and our viscous-Coulomb ansatz (bottom left). \
For Coulomb friction, this function has point discontinuities at those velocities that put any of the feet into static friction (colored dots, color matched across sub-figures). \
For each friction model, we also indicated the body velocity at force-moment balance (red stars).\
The plots demonstrate that the viscous-Coulomb ansatz gives rise to a simple quadratic, whereas Coulomb friction produces an almost everywhere smooth function with point discontinuities, yet both produce very similar solutions.
}}\label{fig:fxy}
\end{figure}

Coulomb friction contacts do not yield a linear relationship between shape-change velocity and body velocity except for the cases where at least three contacts are in static friction and in general configuration. 
Consider the 1-dimensional (1D) case (\refFig{1d_ex}), where a robot constrained to move along the $x$ axis without changing pitch angle, is driven by $M$ legs such that each foot is commanded to move at a potentially different velocity relative to the body $v_i^B$ for $i=1,\cdots,M$.
Further, assume for simplicity that all the feet bear the same normal load $N$.
Denote the friction coefficient as $\mu$, foot velocity in the world frame as $v_i^W$, and the unknown robot body velocity as $v_B$.

\subsubsection{Coulomb friction solution is the median}
The Coulomb friction acting at each foot is $f_i = -\mu N \sgn(v_i^W) = -\mu N \sgn(v_B+v_i^B)$.
At force balance, we want to find $v_B$, such that 
$$
0 = \sum_{i=1}^M -\mu N \sgn(v_B+v_i^B)
$$
Without loss of generality, assume the feet are numbered in a monotone order of foot speed.
In this 1D example force balance is found when half of the feet are slipping in one direction and the other half are slipping in the opposite direction.
If $M$ is odd, assume foot $i$ is moving at the median speed relative to the body. 
If $v_B=-v_i^B$, that foot will be in static friction, and the remaining feet will be in force balance. 
Other solutions are possible, depending on the size of the gap between static and dynamic friction, but the median speed foot gives a solution for any plausible choice of friction coefficients. 
If $M$ is even, assume that the median speed is between $v_i^B$ and $v^B_{i+1}$. 
Choosing $v_B$ anywhere in the (open) interval between these velocities will produce force balance. 
Here additional solutions are also possible if there is a sufficiently large gap between static and dynamic friction coefficients, but the median speeds solution will always exist.

\subsubsection{\expandafter\MakeUppercase\CV\ friction solution is linear}
\Chg{
  Our \CV\ friction ansatz modifies the contact force equations to be linear in the contact velocity vector. 
  In the 1D case, this becomes 
} $f_i = -\mu N v_i^W$.
At force balance,
$$
0 = \sum_{i=1}^M -\mu N (v_B+v_i^B)
$$
The immediate solution is $v_B = -\text{mean}(v_i^B)$ -- a unique linear solution which in many cases is quite close to the median.
The linearity of the \CV\ friction force-balance solution extends to 2D and full 3D, yielding the local connection.

\subsubsection{Static friction in 1D is ill-posed}

Suppose the system moves without any slipping by coordinating the motions of the legs: $v_1^B = v_2^B = \cdots= v_M^B$.
The \CV\ friction model would correctly and robustly estimate the body velocity $v_B = - v_1^B$, whereas the Coulomb friction model would be at a singularity with an ill-posed numerical problem for $|v^W_i|\rightarrow 0$.
Any level of noise in the velocities would change the results. 

\subsection{Contributions}
}

We set out to understand the results of \cite{Walk_slither_Zhao2022} -- why do multiple legs with (presumably) Coulomb friction interactions with the substrate produce a local connection between shape velocity and body velocity which is, algebraically at least, incompatible with such a model.
To address the question through experiment we assembled a hexapedal robot with 6-Degree of Freedom (DoF) force-torque sensors at the base of each leg, enabling the contact forces to be measured directly (see \cite{Wu2023Calib}). 
To address the question through theory, we constructed a highly simplified quasi-static model of body-ground interaction, and replaced the Coulomb friction term which is linear in normal force but non-smooth and non-linear in contact velocity with a friction ansatz that is bilinear in normal force and contact velocity~\cite[Chapter 4]{zhao2021thesis}.

Here we present a refined version of this model and show that it correctly predicts the interplay of forces measured with the robot.
We resolve the seeming contradiction of having the ``wrong'' (ansatz) friction model produce the ``correct'' forces, by showing that Coulomb friction and our ansatz produce very similar motion predictions, for reasons we partially demonstrate in section \ref{sec:discuss} and in \cite{Wu2019friction}.
From a computational perspective, we present a numerical study demonstrating that our computation time is almost independent of the number of legs, unlike the behavior of popular state-of-the-art robot simulation tools. 

\vspace{5mm}
\section{Model algorithm}

We propose an algorithm to estimate world frame body velocity from its body shape and shape-changing velocity at the current time frame.
\Chg{Our algorithm takes as inputs: (1) the positions $q_j$ and velocities $\dot q_j$ of the robot's feet in the body frame; (2) the spring stiffness $k_j$ of each leg; (3) the friction coefficients $\mu_j$ and friction anisotropy $\xyOf{w}{j}$.
As outputs it provides: (1) body height $\zOf{p}{0}$, pitch $\alpha_x$, and roll $\alpha_y$ slopes; (2) body velocities $\xyOf{\dot p}{0}$ and $\dot\theta$; (3) 3D forces at the feet $F_j$.

The model assumes the robot's (1) pitch and roll angles are small; (2) pitch, roll, and height vary slowly, so their derivatives can be assumed to be zero; (3) the system's inertia is irrelevant since the system is always at force-moment balance; (4) the ground is horizontal at the point of contact of each foot.

The algorithm is composed of two steps: (A) find which feet are in contact with the ground and estimate their gravity-induced loading using a spring support model; (B) use force and torque balance to construct the linear local connection model, invert it, and use that to estimate the planar body velocity.
}

The inputs to the spring support model (A) are: 
(1) the 3D positions of the feet $q_j$ in the body frame; (2) the spring stiffness $k_j$ of each leg.
The outputs of the spring support model are: (1) body height $\zOf{p}{0}$, pitch $\alpha_x$, and roll $\alpha_y$ slopes; (2) gravity-induced loading on each foot $\zOf{F}{j}$ and, implicit in that, which feet are in contact with the ground.

Once the contacting feet are known, we solve for force and moment equilibrium using a \CV\ friction ansatz which is bi-linear in $\zOf{F}{j}$ and the foot sliding velocities in the world frame $\xyOf{\dot p}{j}$, providing an local connection model (B). 
The inputs to the connection model are:
(1) the 2D positions $\xyOf{q}{j}$ and velocities $\xyOf{\dot q}{j}$ of the feet in the body frame; (2) the friction coefficients $\mu_j$ and friction anisotropy $\xyOf{w}{j}$; (3) the gravity induced loading $\zOf{F}{j}$ computed in (A).
The outputs of this local connection model are body velocities: (1) body velocities $\xyOf{p}{0}$ and $\dot\theta$; (2) 2D traction forces at the feet $\xyOf{F}{j}$.

Suppose we are given a system with $N$ legs (or other contacts), indexed by $j=1\ldots N$. 
The time-varying foot positions in the body frame of reference are given by $\q_j\in\R^3$, $j=1\ldots N$.
We assume the transformation from body frame to world frame is given by a time-varying rigid body transformation $\tw \in SE(3)$.
The world frame foot positions $\p_j$ and velocities $\dot\p_j$ are

\begin{align}
    \p_j & := \tw\q_j \label{eqn:xfm}\\
    \dot\p_j &= \dot \tw \, \q_j + \tw \dot \q_j
      = \tw \left[ \tw^{-1} \dot \tw \q_j + \dot \q_j \right]
\end{align}
Let $\p_0$ represent the origin of the body frame \Chg{at the CoM}. 
We assume a simplified form for the rigid body transformation approximation $\tw'$, where pitch $\alpha_y$ and roll $\alpha_x$ angles are small, so they can be approximated by their first-order Taylor approximation.
We also assume the rigid body motion is only time-varying in the horizontal plane, i.e. $\alpha_x$ $\alpha_y$ and $\zOf{\p}{0}$ vary so slowly that their derivatives can be approximated by $0$.
\Chg{Specifically, $\tw'$ and $\tw'^{-1}\dot\tw' $ are}

\begin{align}\label{eqn:tw}
    \tw' :&= \begin{bmatrix}
        \cos \theta & -\sin \theta  & \alpha_x & \xOf{\p}{0} \\
        \sin \theta &  \cos \theta  & -\alpha_y & \yOf{\p}{0}\\
-\alpha_x & \alpha_y & 1 & \zOf{\p}{0} \\
        0 & 0 & 0 & 1
    \end{bmatrix}\\
    \tw'^{-1}\dot\tw' :&= \begin{bmatrix}
        0 & -\dot\theta  & 0 & \xOf{\dot\p}{0} \\
        \dot\theta & 0  & 0 & \yOf{\dot\p}{0} \\
0 & 0 & 0 & 0 \\
        0 & 0 & 0 & 0
    \end{bmatrix}.
\end{align}
Because of these simplifying assumptions, we can decouple the movements in the $xy$ plane, and the physical units of vertical and horizontal length are decoupled.
We use the planar rotation $\Rot := \left[ \begin{smallmatrix} \cos\theta& -\sin\theta \\ \sin\theta & \cos\theta \end{smallmatrix} \right]$, and $\Skw := \left[ \begin{smallmatrix} 0& -1 \\ 1 & 0 \end{smallmatrix} \right]$ to represent foot position and velocity in world frame $xy$ plane with:

\begin{align}
    \xyOf{\p}{j} &= \Rot \xyOf{\q}{j} + \xyOf{\p}{0} \\
    \xyOf{\dot\p}{j} &= \Rot \left( \dot \theta \Rot^{-1} \Skw \Rot \xyOf{\q}{j} + \xyOf{\dot\q}{j} \right) \nonumber \\
    		& \qquad + \xyOf{\dot\p}{0} + [\alpha_y, -\alpha_x]^T\zOf{\q}{j}\label{eqn:vxy}
\end{align}

\subsection{Spring Support Model : finding the contacts}\label{sec:contact}

\begin{figure*}[!t]
\centering
\includegraphics[width=\textwidth]{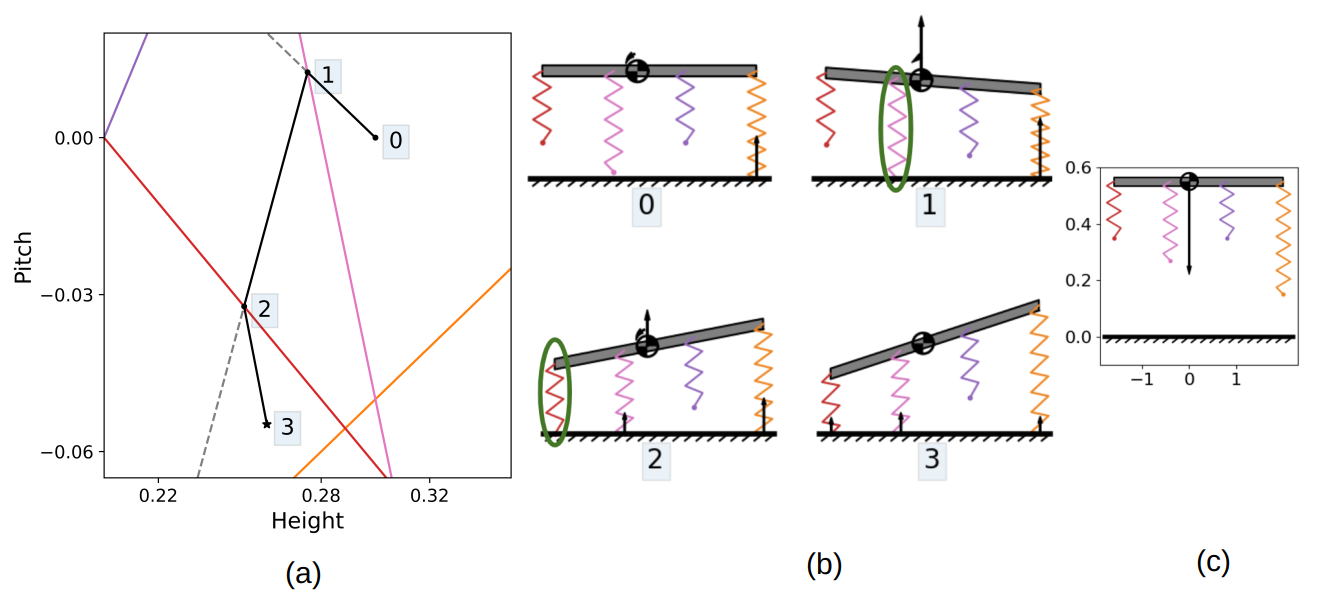}
\caption{%
  Visualization of the search for contact state in a 2D ``robot''. %
  We indicated the height and pitch ($\zOf{\p}{0}$,$\alpha$) states searched (labels 0-3 of (a)), and visualized the pose and contacts of the ``robot'' in each of these states (corresponding labels of plots in (b)). %
  Each ``robot'' leg (zigzag lines in (b)) defines a corresponding codimension 1 plane (line here) in ($z_0$,$\alpha$) at which it contacts the ground (colored lines in (a) with color same as the leg in (b), (c)). %
  At a $\zOf{\p}{0}$ above the plane, the leg is in the air; below it, the leg will be in contact and generate normal forces. %
  With each state being searched (number label in (a)), there is a closed-form solution of the force equilibrium, which we connect to that state with a line interval (black in (a)). %
  If the equilibrium lies in the same contact state the algorithm terminates (star; step 3). %
  Otherwise, the portion of the line segment in another contact state is counter-factual (black dashed in (a)). %
  Instead, we switch to the new contact state and solve it again. %
  Each such transition between contact states lies on a plane corresponding to the leg that made contact (black dot in (a); circled leg in (b)). %
}\label{fig:robot_2d}
\end{figure*}

\Chg{In this section, we show how to decouple the roll, pitch, and vertical (z-axis) motion of the robot, determine which legs are in contact with the ground, and what supporting force each leg generates.}
We model the robot as a ``body plane'', with each leg assumed to be a vertical spring attached to this plane.
We assume the system is at force and moment balance.
A simplified version of this model, without accounting for roll and pitch, can be found in~\cite{revzen-2018-dw-tsml, zhao2021thesis}.
A similar spring-leg model was used to study legged animals and robots in \cite{toppling-table-Usherwood2018} and later in \cite{multilegged-control-Chong2021}, but they did not specify how to determine which legs are in contact.

Consider a pitch, roll, and height state $\s = (\alpha_x, \alpha_y, \zOf{\p}{0})$.
From \eqref{eqn:xfm}, we have 

\begin{equation}\label{eqn:zpj}
    \zOf{\p}{j}(\s) := -\alpha_x \xOf{\q}{j} + \alpha_y \yOf{\q}{j} + \zOf{\q}{j} + \zOf{\p}{0}
\end{equation}

\Chg{Taking $0$ to be the ground level, and up being the positive z-axis direction, legs with $\zOf{\p}{j} < 0$ are in contact with the ground.}
Assuming the normal supporting force $\zOf{F}{j}(\s)$ is linearly dependent on $\zOf{\p}{j}$, we define the individual leg normal force and the resulting planar moment function by,

\begin{align}
    \zOf{F}{j}(\s) & := 
    \left\lbrace \begin{matrix}
         & -K_j\ \zOf{\p}{j}(\s) \quad \text{if }\zOf{\p}{j}(\s)<0 \notag \\\\
		&0 \quad \text{otherwise}
    \end{matrix} \right. \\
    \xOf{M}{j}(\s) & := -\yOf{\q}{j}\zOf{F}{j}(\s) \qquad
    \yOf{M}{j}(\s) := \xOf{\q}{j}\zOf{F}{j}(\s),
\end{align}
and we denote the total force and moment by,

\begin{align}
 F_z(\s) = \sum_{i=1}^N  \zOf{F}{j}(\s), \quad  M_x(\s) = \sum_{i=1}^N  \xOf{M}{j}(\s), \nonumber\\
  M_y(\s) = \sum_{i=1}^N  \yOf{M}{j}(\s).
\end{align}

\Chg{Without loss of generality, we number $\zOf{\p}{j}$ in non-increasing order}. 
When $\alpha_x = \alpha_y = 0$, the total normal force at height $z$ such that $\zOf{\p}{N_k} \leq  -z < \zOf{\p}{N_k+1}$ is $F_z([0,0,z]) = \sum_{j=1}^{N_k} K_j(z + \zOf{\p}{j})$.
\Chg{We scan values of $z$, $z := -\zOf{\p}{N_k}$,} starting with $N_k=1$, where only the lowest foot is in contact.
We increase $N_k$ until $F_z([0,0,-\zOf{\p}{N_k}])\leq Mg < F_z(0,0,-\zOf{\p}{N_k+1})$, and then linearly \Chg{solve using the} slope $K_{N_k+1}$ to find $z^*$ such that force balance is achieved.
For that $z^*$, legs $k=1\ldots N_k$ are in contact with the ground.
Throughout the paper, we use the index $k$ to vary only over legs that in contact with the ground based on this criterion, and by $\zOf{F}{k}$ the normal force of those legs.

Next, we solve for the full state, $\s$, containing the small pitch and roll angles, and the body height, maintaining vertical force balance and moment balance of the moments generated by the normal forces, i.e.  $F_z - Mg= M_x = M_y =0$.
We start with an initial condition $\s_0=(0,0,z^*)$, with $F_z = Mg$.
Taking $\alpha_x$,$\alpha_y$, $z^*$ as unknowns and holding the legs in contact constant, these values are a solution $\s^*_0$ for a 3-dimensional linear system.
We check whether the legs in contact at $\s^*_0$ are the same as in $\s_0$; if so, then $\s^*_0$ is the result of our model. 
If not, then we search along the line segment starting at $\s_0$ and ending at $\s^*_0$ for the first change in contacts, which must therefore occur on a plane describing the contact condition for the first leg which would change contact state going along this line segment.
This transition point is taken as $\s_1$, and the process repeats for the new legs in contact.
Because contact forces are zero on the corresponding contact condition plane, $F_z$, $M_x$, $M_y$ are continuous through the change in contacting legs.
The detailed expression of the equations is in \S\ref{sec:app-normal}.

As the search iterates, we may encounter a state where only one or two legs are in contact, and the linear force torque balance equation becomes under-determined.
To resolve these states we \Chg{use the} additional assumption that the origin of the body plane is the center of mass of the robot body. 
Under this assumption, when there are fewer than three legs in contact, the \Chg{CoM, at a known location,} generates a moment around the contact point(s) and we tilt the body plane, i.e. change $\alpha_x$ and $\alpha_y$, approximating the rotation this moment would induce \Chg{until reaching an angle where} an additional leg contacts the ground. 

We proceed to describe the tilting directions as if they were rotations with an axis. 
However, the actual linear map they describe is a shearing operation whose neutral plane intersects the $xy$ plane on a line containing the rotation axis.
When only one leg is in contact, the rotation is in the plane containing the leg and the COM, around the contact point.
When two legs are in contact, the rotation is around the line connecting their contact points, and in the direction of the moment, the net $F_z$ generates around this line.
The detailed solution is in \S\ref{sec:app-1c} and \S\ref{sec:app-2c}.

We used a 2D ``robot'' in $xz$-plane to visually illustrate our search algorithm in \refFig{robot_2d}.
In this 2D case, the algorithm searches for robot height and pitch using foot position in $x,z$ coordinates. 
The 3D model extends the contact switching lines in 2D searching space to planes in 3D, and its visualization can be found in \refFig{search_3d}.

\subsection{Local connection model : traction forces}\label{sec:planar}
After knowing which legs are in contact and their gravity loading we solve for the body planar velocity $(\xyOf{\dot\p}{0},\dot\theta)$, obtained by imposing force and moment balance.
While classical approaches suggest Coulomb friction is the correct tribological model for sliding dry contacts, we show that a \CV\ ansatz which is bilinear in both loading force and sliding velocity makes for a linear system of equations that leads to a local connection model.

\subsubsection{Friction forces}
The classical approaches to mechanics suggest that the contact between foot and ground should be modeled by Coulomb friction (middle term below), \Chg{defining the use of $\HT_k$}

\begin{align} 
  \xyOf{F}{k} &= -\frac{\xyOf{\dot\p}{k}}{\|\xyOf{\dot\p}{k}\|}\mu_k \zOf{F}{k} = \HT_k \xyOf{\dot\p}{k}. \label{eqn:viscf}  
\end{align}
The choice of $\HT_k = -\mu_k \zOf{F}{k} / {\|\xyOf{\dot\p}{k}\|}$ would provide equality, \Chg{i.e. Coulomb friction}, but this would \Chg{also} produce the well-known problem of singularity at $\xyOf{\dot\p}{k}=0$.
Define $v_k := \|\xyOf{\dot\p}{k}\|$.
We explore the tractability of alternative friction models using

\begin{align}
    \HT_k := -\mu_k \zOf{F}{k} \frac { \varepsilon + v_k }{ \varepsilon + v_k^2 }.\label{eqn:h2}
\end{align}
When $\varepsilon\to 0$, $\HT_k$ \Chg{approaches that of the} Coulomb friction model; when $\varepsilon\to\infty$, $\HT_k\to-\mu_k  \zOf{F}{k}$, the friction force becomes $\xyOf{F}{k} = -\mu_k  \zOf{F}{k}\xyOf{\dot\p}{k}$, a combination of viscous and Coulomb friction, depending \Chg{linearly} on both slipping rate and normal force.

We further deconstruct \eqref{eqn:viscf} in terms of $\dot \theta$, $\xOf{\dot\p}{0}$, and $\yOf{\dot\p}{0}$:

\begin{align}
\xyOf{F}{k} &= \HT_k \left( \Rot \left[ \dot \theta \Rot^{-1} \Skw \Rot \xyOf{\q}{k} + \xyOf{\dot\q}{k} \right] + \xyOf{\dot\p}{0}\right) \notag\\
 &= \left( \HT_k \Skw \Rot \xyOf{\q}{k} \right) \dot\theta + \HT_k\,  \xyOf{\dot\p}{0} + \left( \HT_k \Rot \xyOf{\dot\q}{k} \right)\label{eqn:linFt}
\end{align}

\subsection{Solving for planar body velocity}\label{sec:planar-sol}

From our quasi-static assumption, we have horizontal plane force and moment balance, i.e. $\sum \xOf{F}{k} = \sum \yOf{F}{k} = 0$ and $\sum \zOf{M}{k} = 0$.
From horizontal force balance, using \eqref{eqn:linFt}, we obtain two equations in $\dot \theta$, $\xOf{\dot\p}{0}$, and $\yOf{\dot\p}{0}$

\begin{align}
    0 = \sum_{k=1}^{N_k} \xyOf{F}{k} &= \left( \sum_{k=1}^{N_k} \HT_k \Skw \Rot \xyOf{\q}{k} \right) \dot\theta  + \left( \sum_{k=1}^{N_k} \HT_k  \right) \xyOf{\dot\p}{0} \nonumber \\
    &\qquad + \left( \sum_{k=1}^{N_k} \HT_k \Rot \xyOf{\dot\q}{k} \right)
    \label{eqn:Fxy0}
\end{align}
The moment exerted by a leg is given by:

\begin{align}
    \zOf{M}{k} & = \xyOf{\p}{k}^\T \Skw \xyOf{F}{k} = 
    \left( \xyOf{\p}{k}^\T \Skw \HT_k \Skw \Rot \xyOf{\q}{k} \right) \dot\theta + \nonumber \\
    &\left( \xyOf{\p}{k}^\T \Skw \HT_k \right)  \xyOf{\dot\p}{0} + \left( \xyOf{\p}{k}^\T \Skw \HT_k \Rot \xyOf{\dot\q}{k} \right)
\end{align}
Giving the obvious third and final equation:

\begin{align}
    0 &= \sum_{k=1}^{N_k} \zOf{M}{k} = \left( \sum_{k=1}^{N_k} \xyOf{\p}{k}^\T \Skw \HT_k \Skw \Rot \xyOf{\q}{k} \right) \dot\theta \nonumber\\
   &+ \left( \sum_{k=1}^{N_k} \xyOf{\p}{k}^\T \Skw \HT_k \right)  \xyOf{\dot\p}{0}  + \left(\sum_{k=1}^{N_k} \xyOf{\p}{k}^\T \Skw \HT_k \Rot \xyOf{\dot\q}{k} \right)
    \label{eqn:Mz0}
\end{align}

In the case when $\varepsilon\to\infty$, $\HT_k$ being rate $\xyOf{\dot\p}{k}$ independent, the three force and moment balance equations are linear in the body velocity $\xyOf{\dot\p}{0}$, $\dot\theta$ and foot velocity in body frame $\xyOf{\dot\q}{k}$.
One could solve the system by 3d matrix inversion.
The detailed expression of the solution is derived in \S\ref{sec:app-planar}. 

In addition to classic Coulomb friction and viscous friction, we consider the possibility that $\HT_k$  can be dependent on slipping direction, modeling forces generated by a wheel, skate, claw, or otherwise non-isotropic frictional contact.
We consider an anisotropic viscous friction model, where $\HT_k$ is a symmetric positive semidefinite matrix, $\HT_k(q):=\Rot\HT_{q,k}(q)\Rot^{-1}$ taken to be independent of $\xyOf{\dot\p}{k}$, but (possibly non-linearly) dependent on all elements of $q$.
We assume that each contact is associated with an enhanced traction direction and associated magnitude, expressed in body coordinates as a vector $\xyOf{\w}{k}$, defined as: 
\begin{equation}\label{eqn:anisotropic}
    \HT_k := -\mu_k \zOf{F}{k} \Rot(\mathrm{I}_2 + \xyOf{\w}{k}\xyOf{\w}{k}^\T)\Rot^{-1}
\end{equation}
This changes the circular cross-section of the friction cone into an ellipsoidal one.
Even with this dependence, the equations \eqref{eqn:Mz0} and \eqref{eqn:Fxy0} are still linear in the velocities $\xyOf{\dot\p}{0}$, $\dot\theta$ and $\xyOf{\dot\q}{k}$.
Similar to \S\ref{sec:planar-sol}, body velocity $\xyOf{\dot\p}{0}$, $\dot\theta$ can still be solved linearly with respect to shape changing velocity $\xyOf{\dot\q}{k}$, giving a general form:

\begin{align*}
    \Rot^{-1}\xyOf{\dot\p}{0} &=: \sum_k \xyOf{A}{k}(q) \xyOf{\dot\q}{k}, \\
    \dot\theta &=: -\sum_k A_{\theta,k}(q)\xyOf{\dot\q}{k},
\end{align*}
Where the $A_{\cdot,\cdot}(q)$ matrices \Chg{are} the kinematic term in the reconstruction equation of geometric mechanics.

\section{Experimental validation}\label{sec:results}

\begin{figure*}[!thb]
\centering
\begin{subfigure}[b]{\textwidth}
\centering
\includegraphics[width=0.6\textwidth]{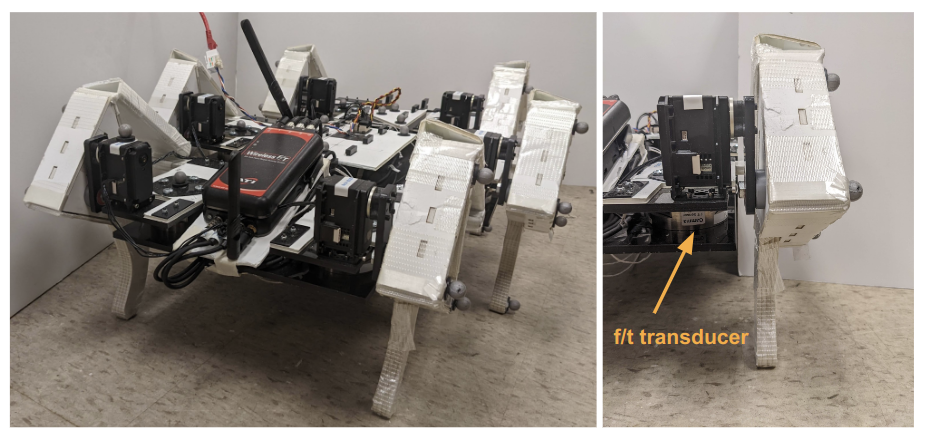}
\end{subfigure}
\vfill
\begin{subfigure}[b]{0.45\textwidth}
\end{subfigure}
\begin{subfigure}[b]{0.45\textwidth}
\end{subfigure}
\vfill
\begin{subfigure}[b]{\textwidth}
\includegraphics[width=\textwidth]{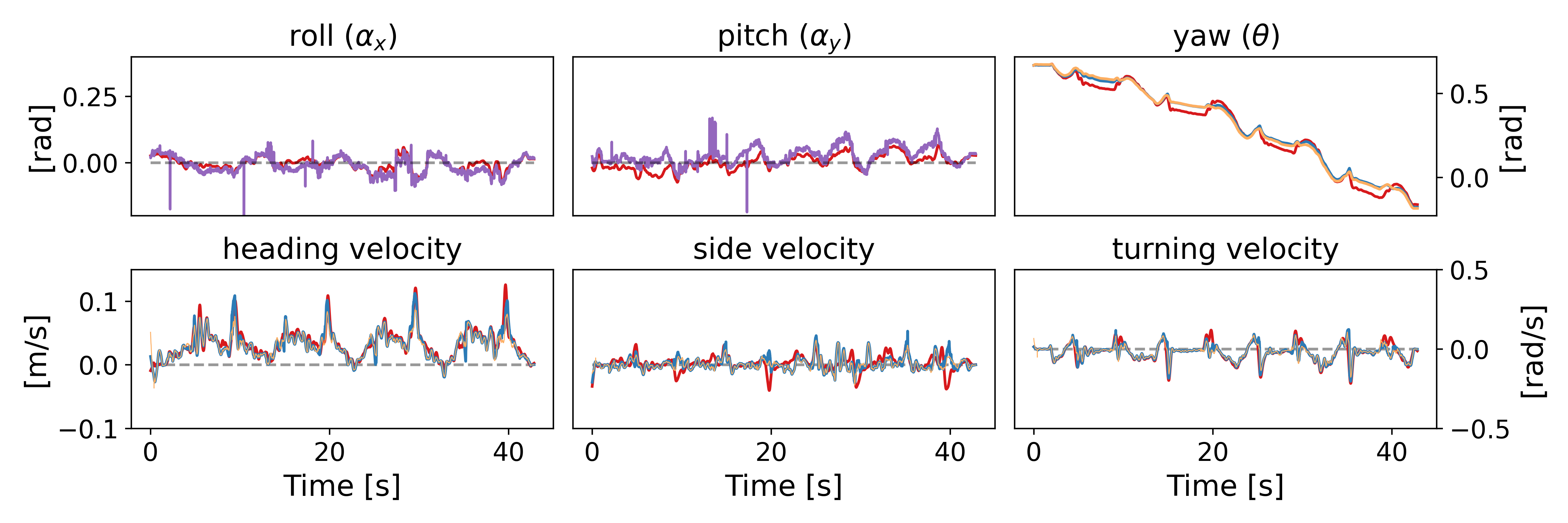}
\end{subfigure}
\vfill
\begin{subfigure}[b]{\textwidth}
\centering
\includegraphics[width=0.85\textwidth]{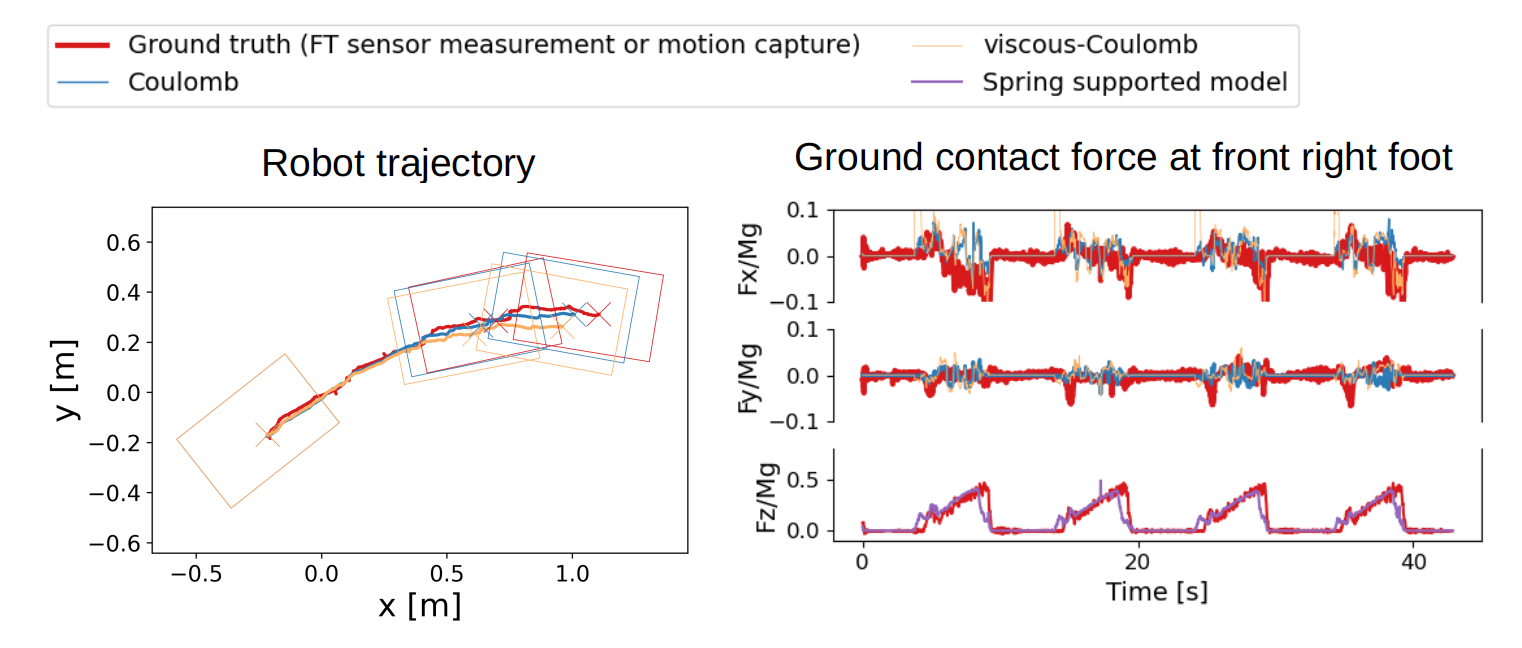}
\end{subfigure}
\vfill
\caption{\Chg{%
  A tripod gait trajectory of BigANT. %
  We plotted the trajectory of the BigANT robot (top row) measured from motion capture (red).\
  We plotted roll, pitch, and yaw angles (second row, left to right), and body velocity components along the robot axis, across the robot, and rotational (third row, left to right). \
  We used motion capture data as ground truth for kinematics (wide red lines). \
  The pitch and roll are the same for both friction models and arise from the spring support model (purple lines), but other outputs are different for Coulomb friction (blue lines) and our \CV\ friction (orange lines). \
  We indicated the body location obtained from motion tracking and by integrating the body velocity predictions (rectangles in ``Robot Trajectory''). \
  We further illustrated the motion by indicating the location of the robot body frame origin (crosses) at the beginning, halfway point, and end of motion (same sub-plot) \
  To compare observed forces to our prediction we plotted details for the Front Right foot: the ground contact force in the $x,y$-axis, and the fraction of supporting force in the $z$-axis. \
  We used force torque sensor measurements as ground truth (same colors and line types).\
  (For all individual foot forces and torques refer to appendix \refFig{tripod-all}.)
  }}\label{fig:FTbigAnt}
\end{figure*}

\begin{figure*}[!t]
\centering
\includegraphics[width=\textwidth]{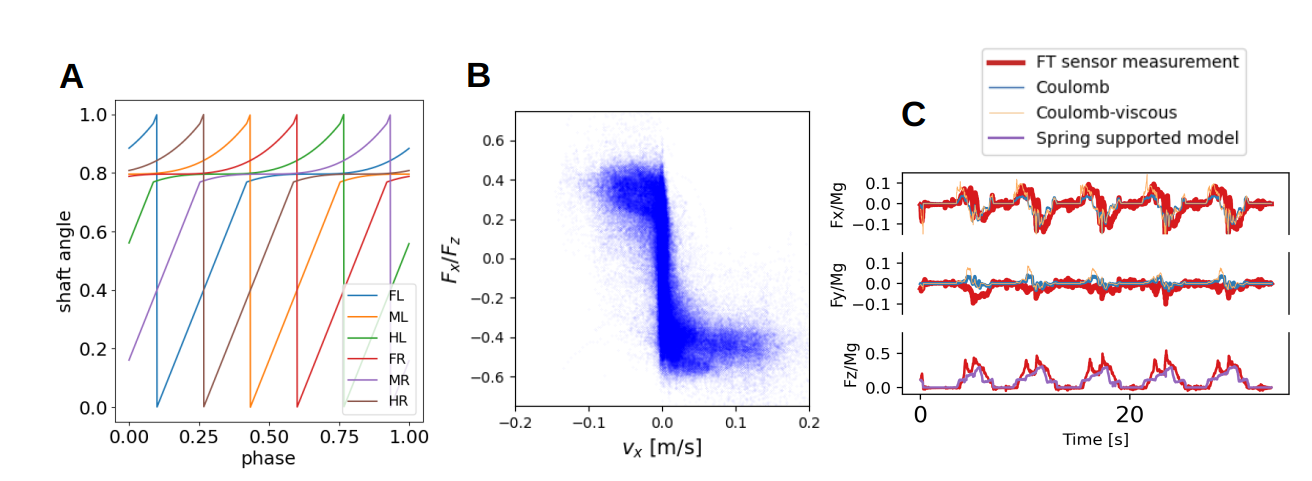}
\caption{%
  We plotted: (A) Metachronal gait phase vs. motor shaft angle for all six legs. %
  (B) Magnitude of slipping velocity vs. magnitude of planar force divided by normal force, overlaying points from all six feet. %
  (C) Ground contact forces \Chg{at the front right foot. For each individual foot refer to appendix \refFig{slipping-all}.} %
}\label{fig:slipping_gait}
\end{figure*}

\Chg{To verify our model's accuracy in practice, we compared our model predictions to ground truth motion capture measurements of three different multilegged robots.}

\Chgp{The marker-based motion capture system used in all experiments consisted of 10 Qualisys Oqus-310+ cameras, running at 100 fps with QTM 2.17 build 4000 software. 
}

\subsection{BigANT with force-torque sensors}

\subsubsection{Robot and measurement setup}
To experimentally verify our algorithm, we built a version of the BigANT robot \Chgp{(\refFig{FTbigAnt})} with a 6-DoF \Chgp{ATI Gamma} force-torque sensor attached to each leg, \Chgp{and used ATI's wireless F/T sensor system to communicate the measurements to the controlling host computer at 100 Hz.}
\Chg{ We calibrated the sensors to report the contact forces at the feet.
This calibration process is of independent interest and published separately in \cite{Wu2023Calib}.}

BigANT is a hexpedal robot that has only one motor \Chgp{(Robotis Dynamixel MX106)} per leg.
The leg drive-train is a four-bar linkage designed to provide advantageous foot clearance and gearing \cite[Chapter 2.2]{zhao2021thesis}.
\Chgp{We manufactured the legs from foamcore (Elmer’s Products Inc. 3/8'' foam board) and fiber reinforced tape (3M Scotch \#8959) using the ``plates and reinforced flexures'' (PARF) technique of~\cite{PARF-Fitzner2017}. 
We laser-cut the base plate for the BigANT chassis from a 1/4'' ABS plastic plate (Mc.Master-Carr).}

We recorded the robot motions with the motion tracking system.
This allowed us to measure the position and orientation of a robot body frame from markers we attached to the robot's chassis, and take the foot positions ($q_j$) relative to that body frame.
We obtained foot velocities $\dot q_j$ by differentiating $q_j$ using a \Chgp{Savitzky-Golay filter (from \texttt{scipy.signal.savgol\_filter}, 2nd order, with window size 25)}.

\subsubsection{Parameter fitting}\label{sec:params}
The remaining inputs to the algorithm were not so easy to determine.
Because our model is quasi-static, the mass plays no direct role, except for its appearance in $Mg$ as the sum total of normal forces at the feet.
The force and moment balance equations remain unchanged regardless of the units selected for force, and these affect only $Mg$, the stiffnesses $K$, and the friction coefficients $\HT$.
We therefore chose $Mg=1$.
Using marker positions, we estimated the robot body's height, pitch and roll according to \S\ref{sec:contact}.

We estimated the spring constants $K$ and two anisotropic friction model coefficients per leg ($\mu_k$ and $\xyOf{\w}{k}$ of \refEqn{anisotropic}) using least-squares minimization of a suitable loss \Chgp{function: we minimized the L2-norm difference between modeled and measured $\zOf{F}{}$ distribution among contacting legs, while adding the variability of $K$ among legs as a regularization penalty.
We assumed anisotropic friction coefficients $\HT_k$ (see \refEqn{anisotropic}), and inferred the parameters $\mu_k$ and $\xyOf{\w}{k}$ for each leg by minimizing L2-norm error between measured forces and forces calculated from slipping velocity measured by motion capture.
We used \texttt{scipy.optimize.least\_squares} for both of these parameter estimation minimizations.}

In total, we fitted 18 constant model parameters to predict a time series of six 3-dimensional leg forces and 6-DoF body velocity measurements, thus there is little risk of over-fitting.

\subsubsection{Coulomb friction solver}
\Chgp{We solved for the Coulomb friction force-balance using \texttt{scipy.optimize.root} with the LM algorithm, hot-starting with the solution of the previous time-step as an initial guess.
Because Coulomb friction is non-smooth, we employed a sequence of smooth approximations to the solution starting from $\varepsilon=10^{-5}$ (from \refEqn{h2}), using each solution as an initial condition for another solution with smaller $\varepsilon$ until the relative change in L2-norm of two consecutive solutions differed less than $10^{-3}$ -- a threshold less than $5\%$ of the median ground speed measured.
In the very rare cases (0.12\% of the BigANT tripod gait dataset) where the Coulomb friction solver failed to converge, we used the ground truth velocity as the initial condition to obtain the Coulomb friction solution; this process resolved the remaining cases.}

\subsubsection{BigANT using tripod gait}\label{sec:tripod}

We first ran the robot through an alternating tripod gait driven with \Chg{the ``Buehler clock'' of \cite{Rhex-Uluc2001}} and the steering strategy described in~\cite{steering-Zhao2020}.
We collected 21 tripod gait trials with the robot running at 0.1Hz \Chg{(dataset at \cite{FT-BigAnt-data})}, with 4-5 cycles in each trial, and a total of 102082 frames consisting of $84\pm 1$ cycles.
The motions of the shaft angles were scheduled to have a slow ground contact phase and a fast leg reset phase. 
We show in \refFig{FTbigAnt} a comparison of forces and kinematics modeled by our multi-contact algorithm with \CV\ friction, our algorithm with classical Coulomb friction, and the experimental measurements.
We integrated body velocity and showed the robot trajectory in \refFig{FTbigAnt}.

Because our physical modeling assumptions only define contact forces up to a positive scale factor, we chose a single positive scalar $\sigma(t)$ for every time step, such that the loss function

\begin{equation*}
\sigma := \arg\min_c \sum_k (c|{\hat F}_k| - |F_k|)^2
\end{equation*}
between the 12 dimensional prediction $\hat F$ and the measured horizontal forces $F$ was minimized.

We reported prediction error statistics in \refFig{err}.
\Chg{The run time for \CV\ friction was 0.19 (0.18, 0.24) [ms/frame; mean, 1st and 3rd quantile].
When running a single approximation with the choice of $\epsilon=10^{-5}$, the Coulomb friction solver took 3.7 (3.1, 3.9) [same].
For a full set of iterations to convergence, Coulomb friction took 10.4 (3.25,15.0) [same], about $\times 54$ slower than our \Chg{\CV\ ansatz based} simulation.

To test \Chg{the} robustness of our modeling approach, we measured the predictions' sensitivity to the model parameters -- the spring constants $K$ and the anisotropic friction coefficients $\mu_k$ and $\xyOf{\w}{k}$ in \refEqn{anisotropic}. 
We compared the prediction using naive parameters with the fitted parameters as described in \S\ref{sec:params}.
We used the same spring constants $K=10$, and only isotropic friction coefficients $\mu=1$ for all legs.
We chose the spring constant magnitude so that the model could lift its feet to perform the appropriate tripod gait.
The body velocity prediction root-mean-square error (RMSE) in the heading direction increased by 14\%, in the side direction decreased by 5.5\%, and turning increased by 1.5\%. 
Roll prediction RMSE decreased by 22\%, pitch RMSE increased by 24\%.
Friction force prediction RMSE among all legs increased by 4\% and 3\% in heading and side direction respectively.
The normal force prediction RMSE among all legs increased by 2\%.
}

\begin{figure*}
\centering
\includegraphics[width=0.85\textwidth]{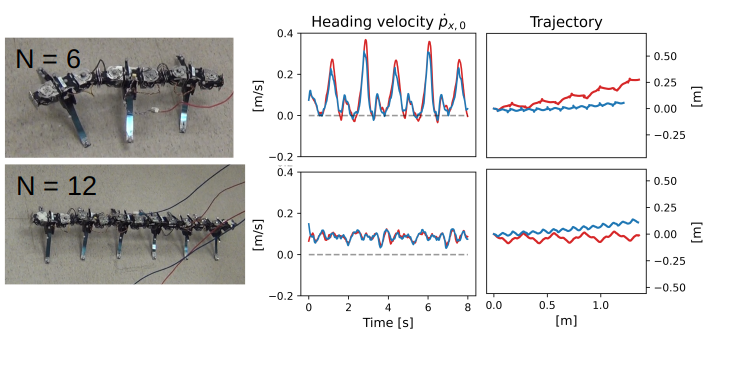}
\caption{\Chg{6 and 12 legged Multipod gait.\
Here we plotted the body velocity and predicted body velocity  at $0.3$Hz and $1.35\,\pi$ phase offset.\
We show Multipods with 6 legs (top) and 12 legs (bottom). \
We plotted the estimated (blue) velocity (middle) and trajectory (right), for comparison with the motion tracking (red).\
Side velocity plots have the same unit and scale as the heading velocity plots.
(For robots with other numbers of legs and more measurements, refer to appendix \refFig{multipod-all}.)\
}}\label{fig:multipod}
\end{figure*}

\begin{figure*}
\centering
\includegraphics[width=0.95\textwidth]{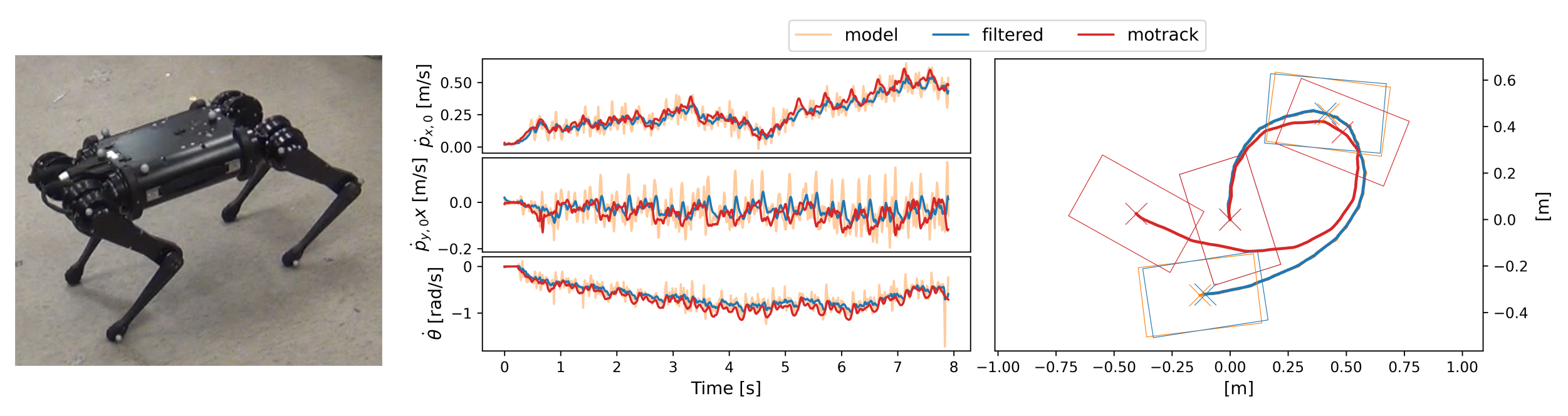}
\caption{\Chg{Modeling of a Ghost Robotics Spirit 40. \
We plotted the estimated velocity (middle, in orange) of the robot (left), filtered estimation (blue) and motion tracking (red) of the absolute motion (right).}
}\label{fig:spirit}
\end{figure*}

\subsubsection{BigANT: metachronal gait}
We wanted to further study why the \CV\ friction model gave similar body velocity and force predictions to those of the classical Coulomb friction model.
Since non-slip motions provide little insight into the question of which friction force model to use, we developed a metachronal gait with exacerbated multi-legged slipping events.
Each foot contacted the ground for $2/3$ of a cycle leading to four feet, two from each side, being in contact with the ground at any time (see \refFig{slipping_gait}). 
To ensure that feet slip, we needed to ensure that the distances between contacting feet change while in stance.
We facilitated this by ensuring that the contact feet have vastly incompatible velocities by choosing the shaft angle to be a cubic function of time during stance. 
We collected 12 metachronal slipping gait trials, with the robot moving forward 4-6 cycles in each. 
In total, the data consisted of 43934 frames and $60\pm 1$ cycles.
The resulting gait produced much more slipping than the tripod gait, with slipping velocities ranging in $(-51.8, 111.8)$[mm/s; 5\% , 95\% percentiles].

To determine whether \CV\ or classical Coulomb friction was indicated in these data, we examined the force measurements from the slipping gait.
Plotting $\xOf{F}{j}/\zOf{F}{j}$ against $\xOf{v}{j}$ (see \refFig{slipping_gait}[B]) shows the expected structure for classical Coulomb friction, namely a step function.

\subsection{Demonstration with other legged systems}
To test whether the proposed model generalizes to other legged systems, we further examined using our model on Multipod robots with 6-12 leaf-spring legs and an undulation gait, and on the commercially available quadruped Ghost Robotics Spirit 40.

\subsubsection{Multipods with 6, 8, 10, and 12 legs}
We used our lab's publicly available Multipod dataset \cite{Multipod-processed-data, Multipod-raw-data, Multipod-raw-videos} used in~\cite{Walk_slither_Zhao2022}.
Each contralateral pair of legs in a multipod has two DoF -- yaw and roll -- and the roll (vertical) motion is further compounded with the spring flexing of the leg itself (see \refFig{multipod}).
\Chgp{To model the body motion of unloaded spring legs, we computed the location of each foot relative to the rigid segment to which its spring was connected based on those motion tracking frames in which the leg was unloaded.
We then extrapolated this unloaded position to the frames where the leg was loaded.}

In \refFig{multipod} we used a slice of these data with the robot running at frequency 0.3Hz and phase offset $1.35\pi$ to demonstrate our algorithm.
We assumed the mass of the robot is linear in the number of legs -- an explicit design feature of these robots -- and set $Mg = N$.
We used $K=1$ as the spring constant and isotropic friction model $\mu_k=1$ on all legs.
\Chg{We provide additional details in \refFig{multipod-all} in the appendix.}

\subsubsection{Ghost Robotics Spirit}
Our physical modeling approach was built upon the assumption that friction dissipates the robot's body momentum quickly in comparison to the time scale of gait motions.
We intentionally selected a commercial quadruped, the Ghost Robotics Spirit, where this assumption breaks down, to test how well the connection-based model could approximate the motion of such a quadruped.
\Chgp{We used the Spirit 40 v1.0 Proto Q-UGV Robot from Ghost Robotics, operated through a Samsung Galaxy A50 with onboard firmware version 0.12.6.}
We collected 921 frames comprising about 9 cycles of motion (see \refFig{spirit}).
Because our model has no inertia, it tends to produce spurious high-frequency changes in its predictions. 
To obtain a more realistic time series, we added a simple model of robot inertia in the form of a first-order IIR lowpass filter $y_n = \gamma y_{n-1} + (1-\gamma) x_n$, where $x_n$ is our raw model prediction and $y_n$ is the filtered prediction.
We manually selected $\gamma=0.15$ to bring the power spectral density (PSD, computed using \texttt{scipy.signal.welch}) of the estimated body velocities close to that of the motion tracking derived velocities.

\subsection{Model runtime analysis}
\begin{figure*}[!hbt]
\centering
\begin{subfigure}[b]{0.45\textwidth}
\includegraphics[width=\textwidth]{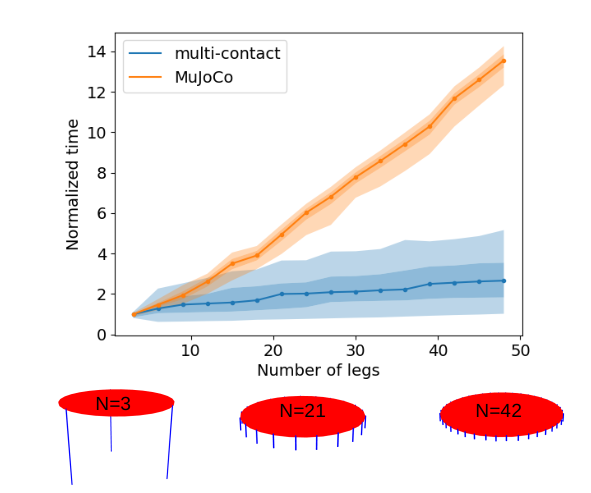}
\end{subfigure}
\begin{subfigure}[b]{0.45\textwidth}
\includegraphics[width=\textwidth]{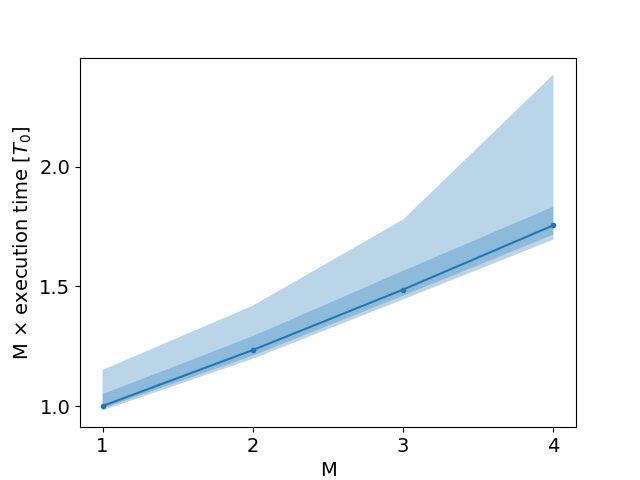}
\end{subfigure}
\caption{%
  (a) Plot of the normalized run time of the multi-contact algorithm and MuJoCo simulation versus the number of legs. %
  We plotted the distribution of time-step computation times on $1000$ randomly initialized configurations for each number of legs from $3$ to $50$. %
  Plot indicates distribution percentiles $2.5$ to $97.5$ (lightly shaded); $25$ to $75$ (shaded); and median (dotted line). %
  The execution times are normalized relative to the median execution time of each simulation on the $3$ leg case. %
  The robot configurations consisted of a disk with $N$ equally spaced legs on the rim as illustrated by examples with $N=3$, $21$ and $42$. %
  (b) Plot of parallelization overhead splitting the algorithm over $M$ threads. %
  The overhead is execution time times $M$ the number of threads, in units of the median execution time on a single thread. %
  Perfectly parallelizable workloads give $1$ whereas unparalellizable workloads give $M$. %
  We plotted the workload distributions at $M=1\ldots 4$ for a hexapod, running $100$ randomized trajectories each $10000$ time-steps long (ribbon with the same quantile as in (a)).
}\label{fig:parallel}\label{fig:run_time}
\end{figure*}

We compared the computation speed between our algorithm using the \CV\ friction model and a widely-used physics simulation engine MuJoCo~\cite[v2.2.1]{Mujoco}.
\Chg{
  MuJoCo performs competitively with other physics simulation engines (see \cite{erez2015simulation}, but also \cite{Yoon-2023-physCmp}), and by restricting itself to convex meshes and relaxing its contact condition, it is often considered for problems with many independent contacts. 
}
Since our focus is on multi-legged contacts, our models consisted of a round disk with 3 to 50 legs equally spaced on its circumference. 
We gave each leg two rotational DOFs, a vertical translation DoF, and limited leg motions so that their workspaces did not overlap.
We tested the execution time of both MuJoCo and our algorithm at 1000 randomly chosen poses and velocities for each number of legs, and re-normalized the running time by dividing by the median execution time of the 3-legged cases, to reveal how each simulation approach scaled with the number of legs (see \refFig{run_time}).
While both algorithms reveal an increase in execution times, our algorithm slows down by less than a factor of 3 with 50 legs, compared with a factor of 13 for MuJoCo.
This suggests that an optimized implementation of our algorithm could be used for multi-legged motion planning for any practical number of contacts.

Because we are using an inertia-free model of physics in the form of a local connection, the body velocity at any instant is only a function of the shape change and shape velocity at that instant. 
Hence, in a homogeneous environment, all time-steps of a motion plan can be computed in parallel.
To demonstrate the performance gains, we simulated 10,000 random poses and velocities of a hexapod robot.
We used $P=1,\cdots,4$ processors to compute the body velocity matrices in parallel, then integrated them in a single linear process (note: this over-estimates the parallelization overhead, since the product of $N$ matrices can be parallelized to take $\log_2 N$ time, but was linear here).
In \refFig{parallel}(b) we show that the algorithm parallelizes well, with the overhead at four processors falling below 1.5, i.e. a net speedup of $4/1.5$.

\section{Discussion}\label{sec:discuss}
Multi-legged robots (with six or more legs) are not widely studied in the robotics community. 
One reason might be that the complexity of modeling the multi-contact ground interaction constrains both motion planning and simulations for design.

Motivated by a previous discovery~\cite{Walk_slither_Zhao2022} -- that multi-legged robots move as if they are governed by a local connection, i.e. quasi-statically -- we developed a simplified ground-interaction model and validated it experimentally.
Our algorithm consists of simulating a spring-supported body in a small-angle approximation for pitch and roll to obtain the vertical foot loadings.
We then introduced the \CV\ ansatz to replace classical Coulomb friction in generating the horizontal forces to produce a linear set of equations that can be solved to give rise to the local connection. 

Our experimental verification demonstrated that while the actual contact forces were, as expected, governed by classical Coulomb friction, our \CV\ friction model gave equally good predictions of both contact forces and body velocities, while computing 50 times faster for a hexapod. 
Our algorithm scales to large numbers of contacts with virtually no change in execution time, and parallelizes with very low overhead.
\Chg{
    Although all our computation was done offline, the computations, even when done in Python, ran about 50 times faster than in real time.
    Thus it is quite clear that our model can easily be used in online form.
}

\begin{figure}
    \centering
    \includegraphics[width=0.4\textwidth]{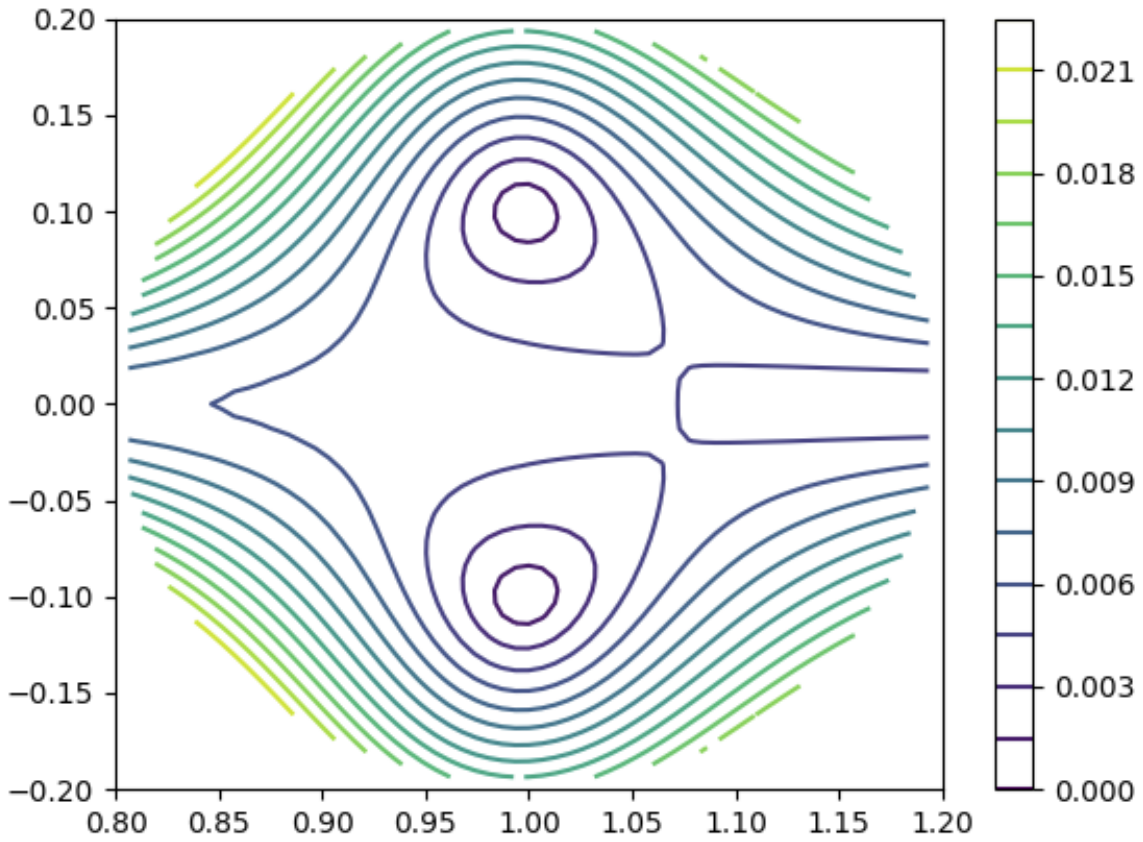}
    \caption{Contour of error between viscous-Coulomb approximation to Coulomb friction around equilibrium velocity. }
    \label{fig:Fv-Fc}
\end{figure}

To understand how a system governed by sliding Coulomb friction can be modeled by a \CV\ friction model, one may examine and compare the relative error of a viscous friction model to that of the ``true'' Coulomb friction.
Because both models are isotropic, we can assume without loss of generality that the velocity is in the $x$ direction.
Because both models are homogeneous, we can assume without loss of generality that the speed is $1$.
What remains is to study the relative error of predicting the Coulomb friction force for contact velocities close to $(1,0)$ and the prediction obtained by using the viscous drag model instead.
In \refFig{Fv-Fc} we present the contours of relative error when using a single viscous friction model instead of Coulomb friction over the specified range of velocities.
The plot demonstrates that with $|\delta v|<0.2|v|$, the \CV\ force prediction for velocity $v+\delta v$ will be within $2\%$ of the classical Coulomb friction force prediction.
The linearity between slipping velocity and friction forces was also observed as an average relationship \Chg{in the numerical simulations of \cite{Wu2019friction} and in  the experiments of \cite{friction-Chong2022}}.

We are thus left with the conclusion that a \CV\ ansatz model for friction produces very similar predictions to those produced by the classical, tribologically accurate Coulomb friction model.
Comparing the motion predictions obtained from both models, they are far more similar to each other than either is to the measured motion, suggesting that the dominant error in these models was not the use of an incorrect friction model.
However, the \CV\ model, in the context of our multi-contact algorithm provides a significant performance boost.
It is faster to compute; it scales better with the number of contacts; and it is easier to parallelize.

From the perspective of physics, that our ansatz produces motion plans as accurate as those produced by Coulomb friction but also provably produces a local connection and principally kinematic motion in the geometric mechanics sense \Chg{is} further justification for the observation of \cite{Walk_slither_Zhao2022} that local connection models \Chg{are} a framework that includes multi-legged locomotion.
While the local connection models of \cite{Walk_slither_Zhao2022} were data-driven, here we have shown that such models can be obtained using a principled modeling approach.

The algorithm we presented here provides merely a starting point -- it is a means for rapidly and accurately estimating multi-contact robot-environment interactions.
Such estimates are building blocks for motion planning, model predictive control, design optimization, and many other potential applications. 
The algorithm itself can be extended to include contacts with non-flat ground.
\Chg{For example, if the ground can be reasonably modeled as horizontal surfaces at various elevations, such as for staircases, we could encompass this in our model with a leg length offset as a function of the foot position in the world frame. 

Modern state observers for quadruped robots use the fusion of proprioceptive information and inertial measurement to estimate state while slipping~\citep{Bloesch2013se}.
These have recently been extended with learning to improve contact estimation and odometry~\citep{Lin2021IKF}.
Like any recursive estimator, such estimators alternate between measurement steps and system evolution steps. 
It would be interesting to explore how to integrate our model as the system evolution operation in such a scheme.
Additionally, the various quantities we estimated by fitting could be converted to slowly varying online estimates, producing an adaptively tuned controller.

Because our proposed model provides a linear relationship between shape space velocity and body velocity, it provides a first-order representation of the dynamics as an affine control system, which could be useful in simplifying the controller design process.

Our model could potentially be used as an extension of the existing models for snake slithering.
First, one might discretize the snake into a 3D shape with appropriately chosen body velocities and stiffnesses.
The simulated snake would then relax into the surface, potentially contacting with only a part of its body, and generate a local connection. 
Perhaps such an approach can provide better models for sidewinding and other fully 3D snake behaviors.
}

We hope that our advances will stimulate the adoption of multi-legged robots in field robotics, provide reliable and adaptable bio-inspired locomotion platforms, and, more generally, enable the modeling of multi-contact mechanics problems.

\subsection*{Acknowledgements} 
We would like to thank Andy Ruina for his key insight in explaining the relationship between our \CV\ ansatz and Coulomb friction.
We also thank the many students who have worked in the BIRDS Lab at the University of Michigan on collecting the large robot motion datasets used herein.

\subsection*{Funding} 
The author disclosed receipt of the following financial support for the research, authorship, and/or publication of this article: This work was supported by the Army Research Office [grant numbers W911NF-17-1-0243, W911NF-17-1-0306]; the National Science Foundation [grant numbers CMMI 1825918, CPS 2038432]; and the D. Dan and Betty Kahn Michigan-Israel Partnership for Research and Education Autonomous Systems Mega-Project.

\subsection*{Conflicting Interests}
The authors declare that there is no conflict of interest. 

\section*{References}
\begin{spacing}{.8}
 \small{
 \setlength{\bibsep}{4.1pt}
\bibliographystyle{IEEEtran}
\bibliography{main}
 }
\end{spacing}

\onecolumn
\section{Appendix}
\subsection{BigAnt with force torque sensors ground contact forces measurement and prediction}
\begin{figure*}[!thb]
\centering
\includegraphics[width=\textwidth]{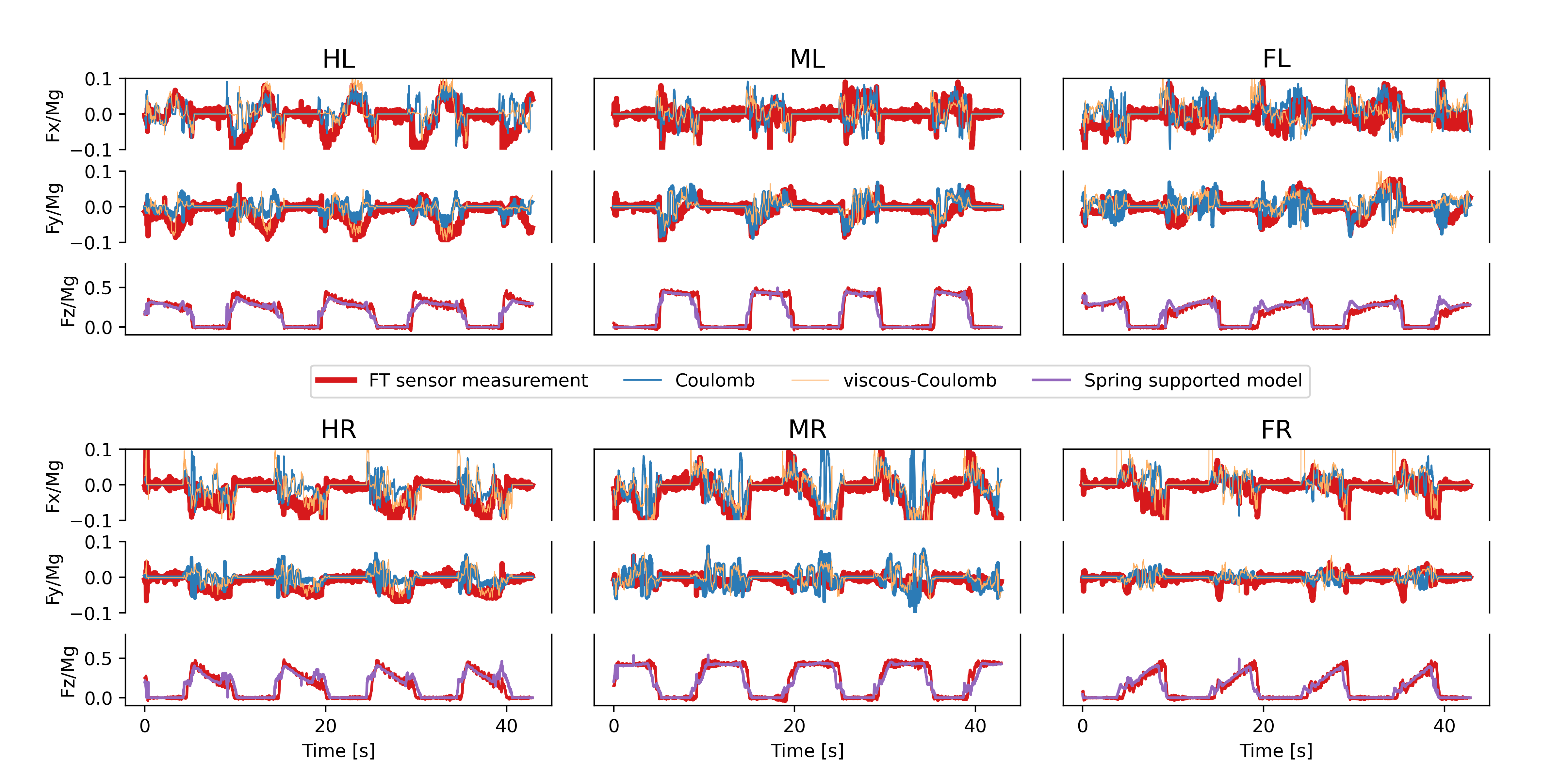}
\caption{%
  Individual leg contact forces measurement and prediction for BigANT robot with tripod gait. %
    We plotted the ground contact force in $x,y$-axis, and supporting force ratio in $z$-axis of each individual leg.\
  We used force torque sensor measurements as ground truth for forces (red).\
We plotted the estimated $z$-axis force ratio (purple), estimated $xy$-axis friction forces (\CV: orange, Coulomb friction: blue).\
(leg names are: HL hind left, ML mid left, FL front left, HR hind right, MR mid right, FR front right)}\label{fig:tripod-all}
\end{figure*}

\begin{figure*}[!thb]
\centering
\includegraphics[width=\textwidth]{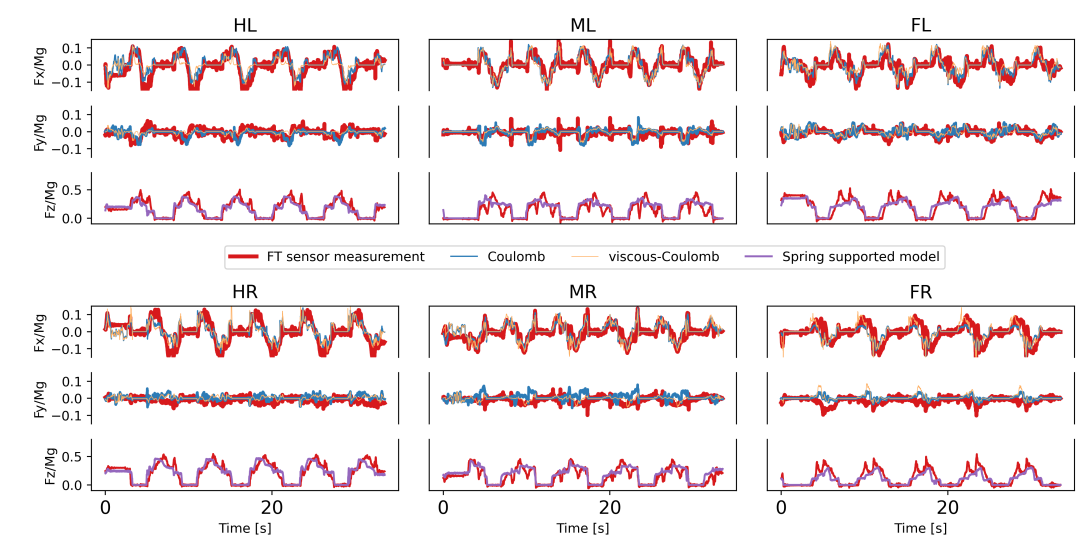}
\caption{%
  Individual leg contact forces measurement and prediction for BigANT robot with metachronal gait. %
    We plotted the ground contact force in $x,y$-axis, and supporting force ratio in $z$-axis of each individual leg.\
  We used force torque sensor measurements as ground truth for forces (red).\
We plotted the estimated $z$-axis force ratio (purple), estimated $xy$-axis friction forces (\CV: orange, Coulomb friction: blue).\
(leg names are: HL hind left, ML mid left, FL front left, HR hind right, MR mid right, FR front right)}\label{fig:slipping-all}
\end{figure*}

\newpage
\subsection{BigANT with force torque sensors modeling error analysis}
\begin{figure*}[!thb]
\centering
\vspace{-0.1in}
\includegraphics[width=\textwidth]{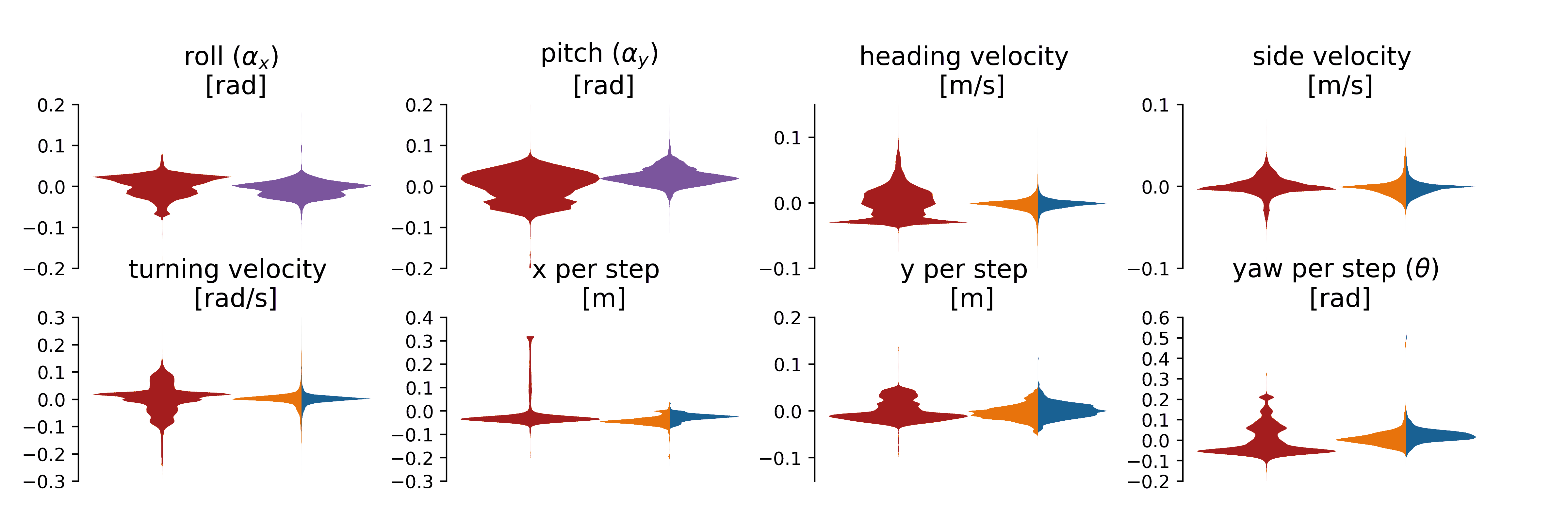}
\vspace{-0.1in}
\includegraphics[width=\textwidth]{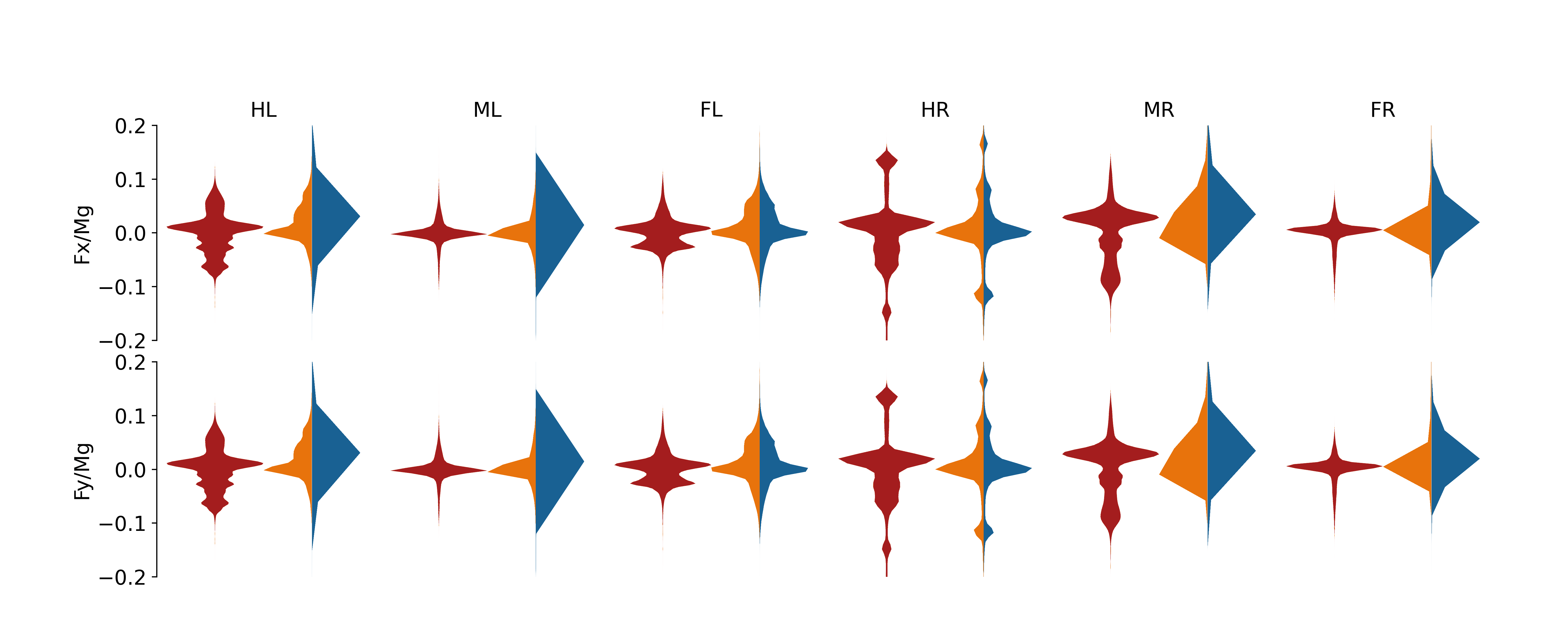}
\caption{%
  Prediction error distributions for BigANT robot with tripod gait. %
  We plotted the distribution of measurement residuals from the mean (red) to compare with residuals of the predictions of the spring supported model (purple), or residuals of predictions from \CV\ (orange) and Coulomb friction (blue). %
  }\label{fig:err}
\end{figure*}

\begin{figure*}[!thb]
\centering
\includegraphics[width=\textwidth]{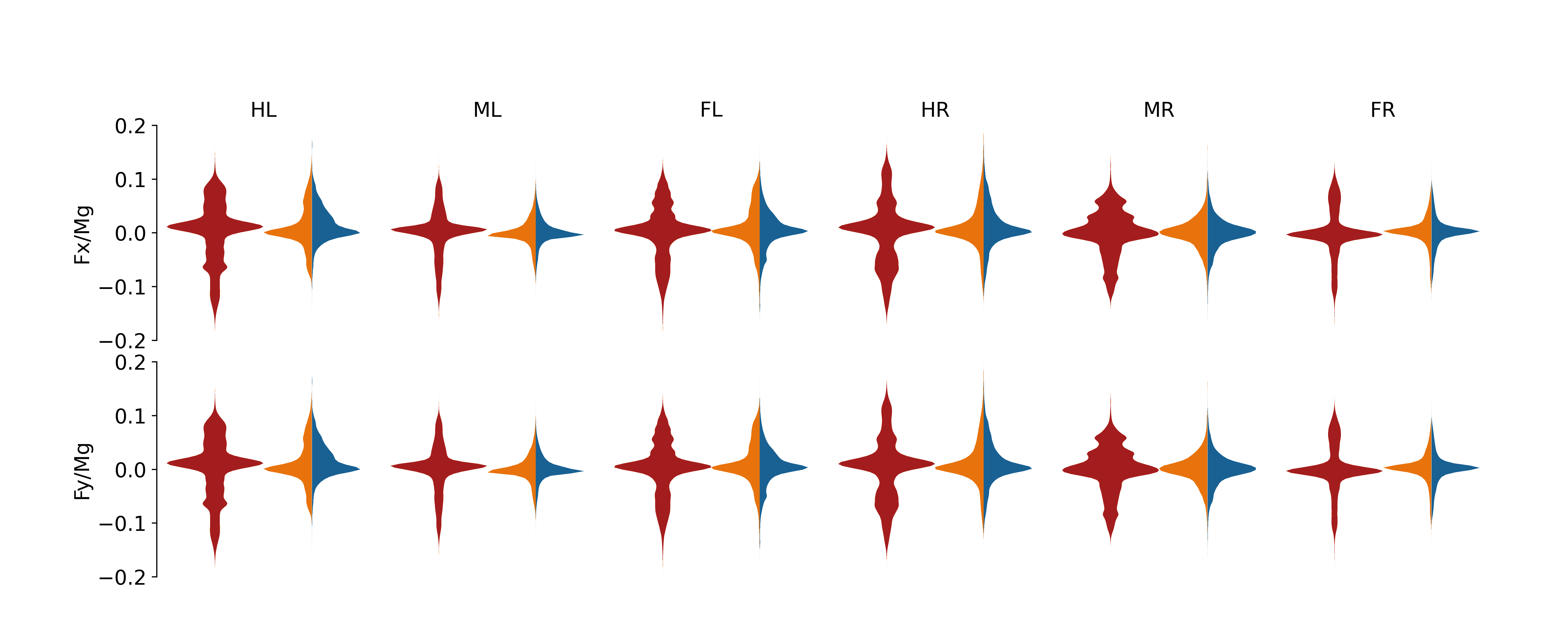}
\caption{%
Planar forces prediction error distributions for BigANT robot with metachronal gait. %
We plotted the planar forces with mean subtracted (red) and model prediction errors for planar forces (\CV: orange, Coulomb:blue). 
}\label{fig:slipping_err}
\end{figure*}

\newpage
\subsection{Multipod}
\begin{figure*}[!thb]
\centering
\includegraphics[width=0.85\textwidth]{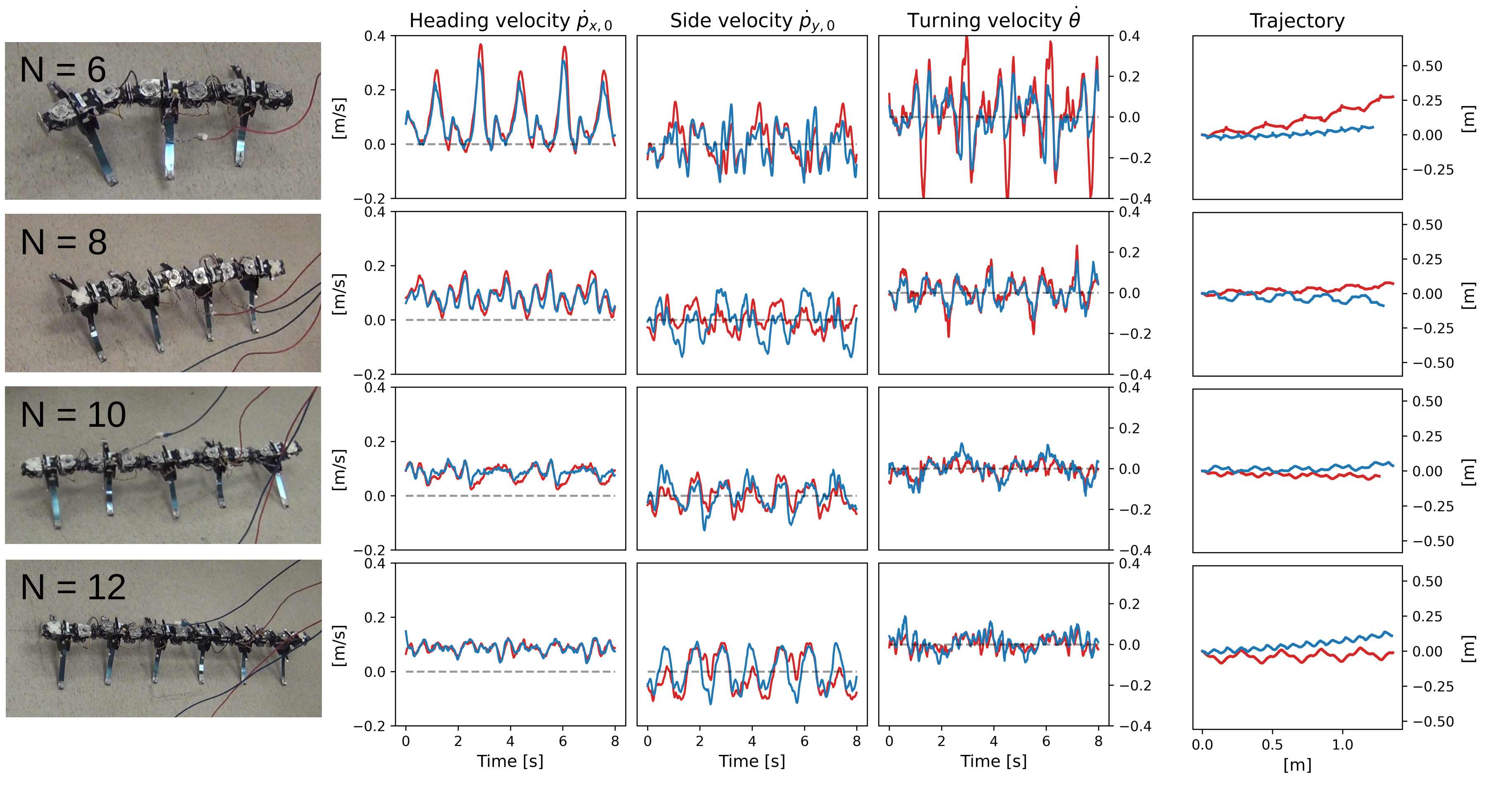}
\caption{Modeling of Multipod with undulation gait at frequency 0.3Hz and phase offset 1.35$\pi$.\
We showed Multipod with 6,8,10,12 legs.\
We plotted the estimated velocity and trajectory (blue), compared with motion tracking (red).\
Side velocity plots have the same unit and scale as the heading velocity plots.}\label{fig:multipod-all}
\end{figure*}

\subsection{Contact detection}
The normal force exerted by a contacting leg:

\begin{align*}
\zOf{F}{k} = K_k(-\alpha_x \xOf{\q}{k} + \alpha_y \yOf{\q}{k} + \zOf{\q}{k} + \zOf{\p}{0})
\end{align*}

The $x,y$ moment resulted from the normal force:

\begin{equation*}
\xOf{M}{k} =  -\yOf{\q}{k}K_k(-\alpha_x \xOf{\q}{k} + \alpha_y \yOf{\q}{k} + \zOf{\q}{k} + \zOf{\p}{0}) \qquad \yOf{M}{k} =  \xOf{\q}{k}K_k(-\alpha_x \xOf{\q}{k} + \alpha_y \yOf{\q}{k} + \zOf{\q}{k} + \zOf{\p}{0})
\end{equation*}

\subsubsection{Normal force and planar moment balance solution}\label{sec:app-normal}
When there are three or more legs in contact, the linear system to solve for normal force and planar moment balance is:

\begin{align*}
    \left[\begin{array}{c}
        \sum_{k=1}^{N_k} \zOf{F}{k}\\
        \sum_{k=1}^{N_k} \xyOf{M}{k}
    \end{array}\right] &= \left[\begin{array}{c}
    -Mg  \\\mathbf{0}
    \end{array}\right]\\
    \sum_{k=1}^{N_k}\left[\begin{array}{c}
    -K_k\xOf{\q}{k}\alpha_x + K_k\yOf{\q}{k}\alpha_y + K_k\zOf{\p}{0} + K_k\zOf{\q}{k} \\
    K_k\xOf{\q}{k}\yOf{\q}{k}\alpha_x - K_k\yOf{\q}{k}^2\alpha_y - K_k\yOf{\q}{k}\zOf{\p}{0} - K_k\yOf{\q}{k}\zOf{\q}{k} \\
    -K_k\xOf{\q}{k}^2\alpha_x + K_k\xOf{\q}{k}\yOf{\q}{k}\alpha_y + K_k\xOf{\q}{k}\zOf{\p}{0} + K_k\xOf{\q}{k}\zOf{\q}{k}
    \end{array}\right]&= \left[\begin{array}{c}
    -Mg  \\0\\0
    \end{array}\right]\\
    \sum_{k=1}^{N_k}\left[\begin{array}{ccc}
    -K_k\xOf{\q}{k} & K_k\yOf{\q}{k} & K_k  \\
    -K_k\xOf{\q}{k}\yOf{\q}{k} & K_k\yOf{\q}{k}^2 & -K_k\yOf{\q}{k}  \\
    -K_k\xOf{\q}{k}^2 & K_k\xOf{\q}{k}\yOf{\q}{k} & K_k\xOf{\q}{k} 
    \end{array}\right]\left[
    \begin{array}{c}
    \alpha_x \\ \alpha_y \\ \zOf{\p}{0}
    \end{array}
    \right]
    &= \left[\begin{array}{c}
    -Mg - K_k\zOf{\q}{k} \\ K_k\yOf{\q}{k}\zOf{\q}{k}\\ -K_k\xOf{\q}{k}\zOf{\q}{k}
    \end{array}\right]
\end{align*}

Let the previous state to be $\s_0 = [\alpha_x^0, \alpha_y^0, \zOf{\p}{0}^0]$ and current solution $\s_1 = [\alpha_x^1, \alpha_y^1, \zOf{\p}{0}^1]$, the next state is obtained by $\s_2 = \s_0 + t(s_1-s_0)$, where $t$ is the minimum solution among all $t_j$ for which $t_j\in[0,1]$. 
Each $t_j$ is calculated from: 

\begin{equation*}
\zOf{F}{j}(\s_2) = K_j(-((1-t_j)\alpha_x^0 + t\alpha_x^1) \xOf{\q}{j} + ((1-t_j)\alpha_y^0 + t_j\alpha_y^1) \yOf{\q}{j} + \zOf{\q}{j} + (1-t_j)\zOf{\p}{0}^0 + t_j\zOf{\p}{0}^1)) = 0
\end{equation*}
Therefore, 

\begin{equation*}
t_j =\frac{\alpha_x^0\xOf{\q}{j} - \alpha_y^0\yOf{\q}{j} - \zOf{\q}{j} - \zOf{\p}{0}^0}{-(\alpha_x^1-\alpha_x^0)\xOf{\q}{j} + (\alpha_y^1-\alpha_y^0)\yOf{\q}{j} + (\zOf{\p}{0}^1 - \zOf{\p}{0}^0)},\qquad t = \min\{t_1, \cdots t_N\}
\end{equation*}

\vfill

\begin{figure}
\centering
\includegraphics[width=0.8\textwidth]{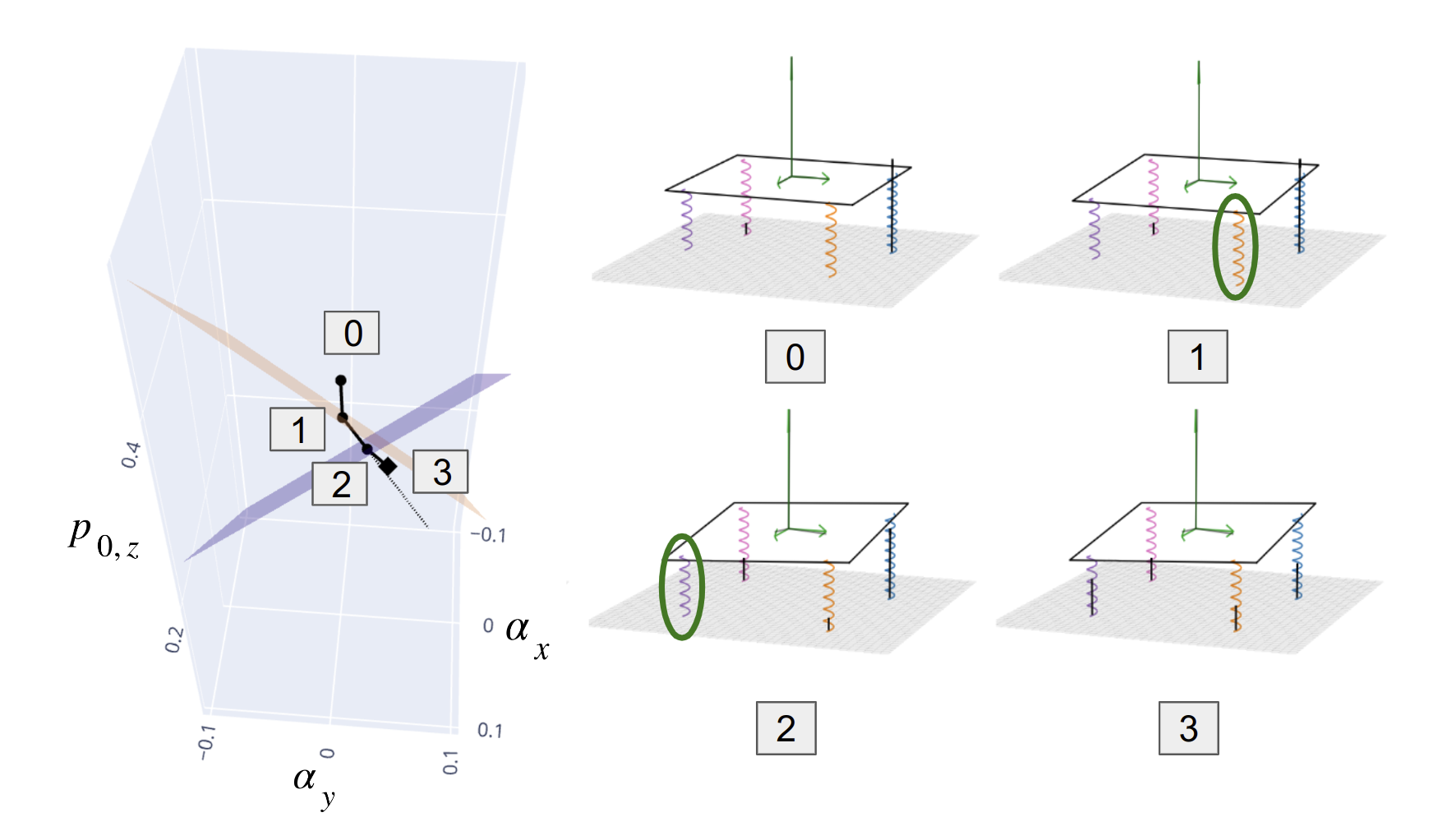}
\caption{Searching iterations for a 3D robot.\
The iterations labeled 0-3 in the search space (left) correspond to the labels below the robot (right).\
The colored surfaces are the contact switching surfaces for legs that has active contact switches in this iteration (the color of a surface corresponds to the color of a leg on the robot).\
It starts with the initial condition, where z-axis force balance is satisfied.\
Then it solves force and moment balance for pitch, roll and height, where the grey dashed line is the direction of the solution given current contact indices.\
It goes along that direction, and stops at the first intersection with a color line, where a leg switches its contact state.\
The iteration continues, until a solution is found within its current polygon, i.e. contact indices do not change.\
The green circle highlights the legs switching contact at that iteration. 
}\label{fig:search_3d}
\end{figure}

\subsubsection{Special case: one leg in contact}\label{sec:app-1c}
When there is only one leg in contact with the ground, denoting it with index 1, we rotate the body frame about the contacting point along the normal direction connecting the contacting leg and origin. 
The direction of rotation is the direction of the torque resulted by robot gravity at the contacting point. 
The direction towards next state is $s_1 - s_0 = [-\yOf{\p}{1}^2-\xOf{\p}{1}^2, \yOf{\p}{1}, -\xOf{\p}{1}]$.
The magnitude of state change $t$ is the minimum amount that results in another contact leg, which can be solved the same as previous section. 

\subsubsection{Special case: two legs in contact}\label{sec:app-2c}
When there are only two legs in contact with the ground, denoting it with index 1 and 2, we rotate the body frame about the line connecting these two contact points.
The direction of rotation is the direction of the torque resulted by robot gravity about the line. 
Therefore, let $a=\yOf{\p}{1} - \yOf{\p}{2}, b=\xOf{\p}{2} - \xOf{\p}{1}, c = \yOf{\p}{1}\xOf{\p}{2} - \yOf{\p}{2}\xOf{\p}{1}$, the direction of the change in pitch and roll becomes $n = [-ca/\sqrt{c^2a^2+c^2b^2}, cb/\sqrt{c^2a^2+c^2b^2}]$.
Together with the change in height, the direction towards next state is:

$$
s_1 - s_0 = [(-n_0^2 - n_1^2)|c|/\sqrt{(a^2+b^2)}, n_0, n_1]
$$
The magnitude of state change $t$ is again the minimum amount that results in another contact leg, which can be solved the same as previous section. 

\newpage
\subsection{Planar motion}\label{sec:app-planar}
The planar force exerted by a leg:

\begin{align*}
\xyOf{F}{k} &= \HT_k \left( \Rot \left[ \dot \theta \Rot^{-1} \Skw \Rot \xyOf{\q}{k} + \xyOf{\dot\q}{k} \right] + \xyOf{\dot\p}{0}\right) \\
 &= \left( \HT_k \Skw \Rot \xyOf{\q}{k} \right) \dot\theta + \HT_k\,  \xyOf{\dot\p}{0} + \left( \HT_k \Rot \xyOf{\dot\q}{k} \right)\\
 &= \left( \Rot\HT_{q,k} \Skw \xyOf{\q}{k} \right) \dot\theta +
 \Rot\HT_{q,k}\Rot^{-1}\,  \xyOf{\dot\p}{0} + 
 \left( \Rot\HT_{q,k} \xyOf{\dot\q}{k} \right)\\
\end{align*}

The $z-$moment exerted by a leg: 

\begin{align*}
    \zOf{M}{k} &:
    = (\xyOf{\p}{k}-\xyOf{\p}{0})^\T \Skw \xyOf{F}{k}\\
    &= 
    \left( (\xyOf{\p}{k}-\xyOf{\p}{0})^\T \Skw \HT_k \Skw \Rot \xyOf{\q}{k} \right) \dot\theta + \left( (\xyOf{\p}{k}-\xyOf{\p}{0})^\T \Skw \HT_k \right)  \xyOf{\dot\p}{0}  + \left( (\xyOf{\p}{k}-\xyOf{\p}{0})^\T \Skw \HT_k \Rot \xyOf{\dot\q}{k} \right)\\
    &= \left( (\xyOf{\p}{k}-\xyOf{\p}{0})^\T \Skw \Rot\HT_{q,k}\Skw \xyOf{\q}{k} \right) \dot\theta + \left( (\xyOf{\p}{k}-\xyOf{\p}{0})^\T \Skw \Rot\HT_{q,k}\Rot^{-1} \right)  \xyOf{\dot\p}{0}  \\ &\quad +\left( (\xyOf{\p}{k}-\xyOf{\p}{0})^\T \Skw \Rot\HT_{q,k} \xyOf{\dot\q}{k} \right)\\
    &=     \left( \xyOf{\q}{j}^\T \Skw\HT_{q,k}\Skw \xyOf{\q}{k} \right) \dot\theta + \left( \xyOf{\q}{j}^\T \Skw \HT_{q,k}\Rot^{-1} \right)  \xyOf{\dot\p}{0} + \left( \xyOf{\q}{j}^\T \Skw \HT_{q,k} \xyOf{\dot\q}{k} \right)
    \end{align*}
    
The linear system to solve for planar force and moment balance is:

\begin{align*}
    \left[\begin{array}{ccc}
        \sum_{k=1}^{N_k} \xyOf{F}{k}\\
        \sum_{k=1}^{N_k} \zOf{M}{k}
    \end{array}\right] &= 0\\
    \sum_{k=1}^{N_k}\left[\begin{array}{cc}
        \Rot\HT_{q,k}\Rot^{-1} & \Rot\HT_{q,k} \Skw \xyOf{\q}{k} \\
        \xyOf{\q}{j}^\T \Skw \HT_{q,k}\Rot^{-1} & \xyOf{\q}{j}^\T \Skw\HT_{q,k}\Skw \xyOf{\q}{k}
    \end{array}
    \right]
    \left[\begin{array}{c}
        \xyOf{\dot\p}{0} \\
        \dot\theta
    \end{array}
    \right] &= 
    \sum_{k=1}^{N_k}\left[\begin{array}{c}
        \Rot\HT_{q,k} \xyOf{\dot\q}{k} \\
        \xyOf{\q}{j}^\T \Skw \HT_{q,k} \xyOf{\dot\q}{k}
    \end{array}
    \right]\\
    \left[
    \begin{array}{cc}
        \Rot & 0 \\
        0 & 1
    \end{array}
    \right]
    \sum_{k=1}^{N_k}\left[\begin{array}{cc}
        \HT_{q,k} & \HT_{q,k} \Skw \xyOf{\q}{k} \\
        \xyOf{\q}{j}^\T \Skw \HT_{q,k} & \xyOf{\q}{j}^\T \Skw\HT_{q,k}\Skw \xyOf{\q}{k}
    \end{array}\right]
    \left[\begin{array}{c}
        \Rot^{-1}\xyOf{\dot\p}{0} \\
        \dot\theta
    \end{array}
    \right] &= 
    \left[
    \begin{array}{cc}
        \Rot & 0 \\
        0 & 1
    \end{array}
    \right]
    \sum_{k=1}^{N_k}\left[\begin{array}{c}
        \HT_{q,k} \xyOf{\dot\q}{k} \\
        \xyOf{\q}{j}^\T \Skw \HT_{q,k} \xyOf{\dot\q}{k}
    \end{array}
    \right]\\
    \left[\begin{array}{c}
        \Rot^{-1}\xyOf{\dot\p}{0} \\
        \dot\theta
    \end{array}
    \right] = \left(  \sum_{k=1}^{N_k}\left[\begin{array}{cc}
        \HT_{q,k} & \HT_{q,k} \Skw \xyOf{\q}{k} \\
        \xyOf{\q}{j}^\T \Skw \HT_{q,k} & \xyOf{\q}{j}^\T \Skw\HT_{q,k}\Skw \xyOf{\q}{k}
    \end{array}\right]\right)^{-1}\sum_{k=1}^{N_k}&
    \left[\begin{array}{c}
        \HT_{q,k}  \\
        \xyOf{\q}{j}^\T \Skw \HT_{q,k}
    \end{array}
    \right]\xyOf{\dot\q}{k}
\end{align*}

\end{document}